\pdfoutput=1
\documentclass[11pt]{article}

\usepackage[final]{acl}

\usepackage{times}
\usepackage{multirow}
\usepackage{latexsym}
\usepackage{amsmath}
\usepackage{graphicx}
\usepackage{algorithm}
\usepackage{algorithmicx}
\usepackage{algpseudocode}
\usepackage{hyperref}
\usepackage{makecell}
\usepackage{caption}
\usepackage{subfigure}
\usepackage[T1]{fontenc}
\usepackage{makecell}
\usepackage[utf8]{inputenc}

\usepackage{microtype}

\title{Generation Meets Verification: Accelerating Large Language Model Inference with Smart Parallel Auto-Correct Decoding}

\author{
Hanling Yi$^1$, Feng Lin$^{1,2}$, Hongbin Li$^1$, Peiyang Ning$^1$, Xiaotian Yu$^1$, Rong Xiao$^1$ \\
$^1$Intellifusion Inc.\\
$^2$Harbin Institute of Technology, Shenzhen\\
\texttt{\{hanling.cuhk,lee.blingner,xiaotianyu.ac,rongxiao\}@gmail.com} \\
\texttt{lin1993@mail.ustc.edu.cn}
}

\begin{document}
\maketitle
\begin{abstract}
%This research endeavors to accelerate the inference speed of large language models with billions of parameters. Our proposed novel method, termed \textbf{S}mart \textbf{P}arallel \textbf{A}uto-\textbf{C}orrect D\textbf{e}coding (SPACE), equips LLMs with the capability to simultaneously generate new candidate tokens and verify previously generated ones. SPACE pioneers two primary innovations: firstly, a semi-autoregressive supervised fine-tuning scheme that enables LLMs to speculatively anticipate several upcoming tokens; secondly, an intricate auto-correct decoding algorithm designed to attain a lossless acceleration of LLM inference speeds. Experimental results on various LLMs show SPACE can achieve 1.5x-4x speedup while provably preserving output quality. 

This research aims to accelerate the inference speed of large language models (LLMs) with billions of parameters. We propose \textbf{S}mart \textbf{P}arallel \textbf{A}uto-\textbf{C}orrect d\textbf{E}coding (SPACE), an approach designed for achieving lossless acceleration of LLMs. By integrating semi-autoregressive inference and speculative decoding capabilities, SPACE uniquely enables autoregressive LLMs to parallelize token generation and verification. This is realized through a specialized semi-autoregressive supervised fine-tuning process that equips existing LLMs with the ability to simultaneously predict multiple tokens. Additionally, an auto-correct decoding algorithm facilitates the simultaneous generation and verification of token sequences within a single model invocation. Through extensive experiments on a range of LLMs, SPACE has demonstrated inference speedup ranging from 2.7x-4.0x on HumanEval-X while maintaining output quality. Code is released at \url{https://github.com/cteant/SPACE. }

\end{abstract}

\section{Introduction}
The majority of large language models (LLMs), including prominent examples like ChatGPT~\cite{brown2020language} and LLaMA~\cite{touvron2023llama}, are autoregressive (AR) in nature. During the inference stage, these AR models generate tokens one by one in a sequential manner. This sequential approach limits parallelism, leading to underutilization of modern parallel computing resources such as GPUs. Consequently, the inference stage becomes memory-bound and the inference latency increases noticeably, particularly with advanced LLMs boasting billions of parameters.
%, where speed is crucial but hindered by the sequential token generation process.
%leading to underutilization of modern parallel computing resources such as GPUs. 
% Consequently, there is a noticeable increase in latency during the inference stage. This issue becomes more pronounced when dealing with advanced LLMs, typically equipped with billions of parameters, where speed is crucial but hindered by the sequential token generation mechanism.

A straightforward method to mitigate the latency is to adapt the model to predict multiple future tokens in parallel. Such models are commonly referred to as semi-autoregressive (SAR) models~\cite{wang2018semi}. 
Nonetheless, the vast majority of LLMs are inherently AR and, hence, unable to perform inference in a SAR manner.  In addition, SAR models commonly experience a deterioration in the output quality due to their parallel decoding nature~\cite{xiao2023survey}. Furthermore, it is worth mentioning that pretraining any LLM from scratch is computationally expensive.
%Besides, SAR models often suffer from a degradation in model quality due to their parallel decoding nature~\cite{xiao2023survey}, not to mention that pretraining a SAR LLM from scratch is computationally expensive.

Another effective way to speed up AR sampling is speculative decoding~\cite{google2023fast, deepmind2023accelerating, miao2023specinfer}. Speculative decoding typically adheres to the `draft-then-verify' paradigm, wherein multiple candidate tokens are initially generated by fast-to-infer smaller models, and are subsequently validated in parallel by the larger LLM. This validation process, based on rejection sampling, ensures that the final output is consistent with the LLM's distribution, thereby achieving lossless speedup. 
Nonetheless, speculative decoding is contingent on the availability of smaller models, which must utilize the same tokenizer as the larger model to function properly. 
%Further, these additional models incur extra memory overhead during inference.
% Nonetheless, speculative decoding requires small model share the same tokenizer with large model and thus limits its adoption.

%For instance,  Stern et al. ~\cite{stern2018blockwise} inserted multiple feedforward heads on top of the Transformer decoder to generate multiple candidate tokens concurrently and used the original head to verify these outputs. Leviathan et al.~\cite{google2023fast} deployed a small model to sample a sequence of tokens, which were subsequently scored by the LLM in parallel.  Miao et al.~\cite{miao2023specinfer} employed multiple small models for candidate tokens generation and then proposed a token-tree verification algorithm for efficient verification of these tokens. 
% However, to the best of our knowledge, all previously established methods either rely on extra smaller models or additional feedforward heads, which consequently results in increased memory consumption during the inference stage.

% To alleviate the memory usage and boost the decoding speed during inference, this paper advocates for the utilization of a semi-autoregressive (SAR) approach. Typically a SAR model can independently generate multiple target tokens in parallel, eliminating the need for supplementary small models or additional trainable parameters. 
%The integration of SAR inference with speculative decoding presents an promising approach for accelerating language model inference. 
Integrating SAR inference with speculative decoding presents a promising approach to accelerate language model inference.
By adapting a model to autonomously generate and validate a sequence of future tokens, we establish an efficient and self-reliant process that greatly enhances the speed of inference. This union yields substantial practical benefits: it eliminates the requirement for smaller auxiliary models, thereby simplifying the overall implementation and reducing memory overhead during inference.
Furthermore, by shifting the emphasis away from precise prediction of multiple tokens towards speculative generation followed by verification, the difficulty of the SAR training phase can be significantly reduced.

%We demonstrate that it is possible to configure a pretrained AR language model to generate multiple future tokens in parallel by employing supervised fine-tuning (SFT) with a manageable dataset of tens of thousands of samples. This process eliminates the need for auxiliary smaller models and keeps the training overhead to an acceptable level. Furthermore, we incorporate the draft-then-verify technique and put forth an efficient inference algorithm. This novel algorithm enables simultaneous generation and validation of candidate tokens through a singular invocation of the model, thereby streamlining the inference process.

%However, the quality of generation in a SAR model often suffers due to parallel decoding, as compared to the AR counterpart. Interestingly, we observe that a SAR language model, if infers in an AR manner, incurs minimal quality loss. 
% Although SAR models typically experience a degradation in the quality of generation due to their parallel decoding nature, our findings reveal a minimal quality loss when a SAR model performs inference in an AR fashion. 
% To speed up the AR inference of the model, we adopt the draft-then-verify paradigm and introduce an efficient inference algorithm. The inference algorithm allows the model to concurrently generate and validate candidate tokens via one model invocation. 

\begin{figure*}[thb]
\centering
\includegraphics[width=5.7in]{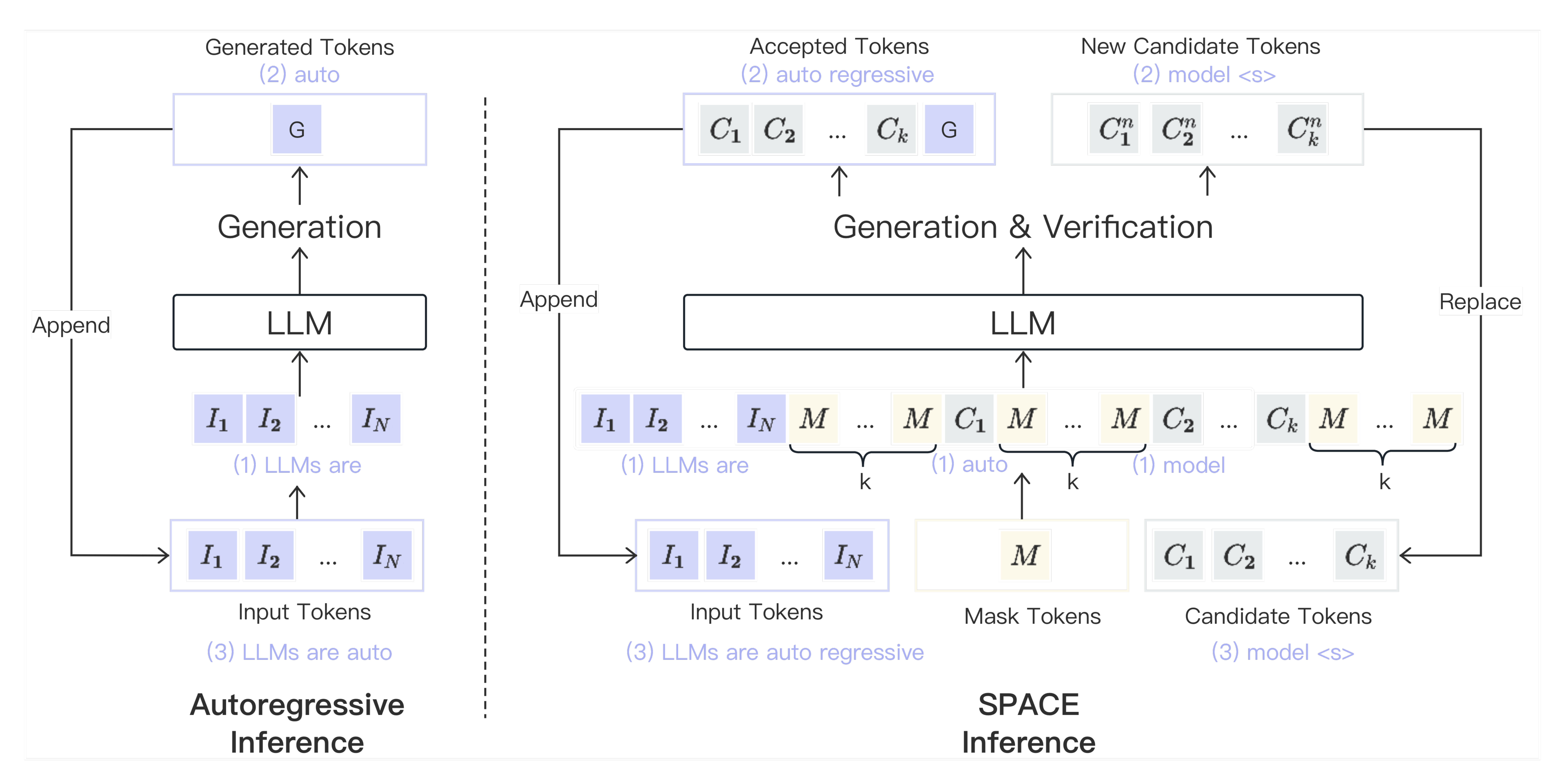}
%\caption{A visual comparison between conventional AR inference (left) and SPACE inference (right) is illustrated. In AR inference, token generation proceeds in a sequential manner, with only one token output per decoding step. In SPACE inference, the input token sequence is augmented with $k+1$ groups of mask tokens and $k$ candidate tokens. The candidate tokens undergo verification, and $k$ new candidate tokens are generated from one of the mask groups after a single model invocation. SPACE allows for a variable number of tokens to be generated in each step, with the quantity ranging from a minimum of 1 to a maximum of $k+1$.}
\caption{A visual comparison between conventional AR inference (left) and SPACE inference (right) is illustrated. In AR inference, token generation proceeds in a sequential manner, with only one token output per decoding step. In SPACE inference, the input token sequence (i.e., ``LLMs are'') is augmented with $k+1$ groups of mask tokens and $k$ candidate tokens (i.e., ``auto'' and ``model''). The candidate tokens undergo verification to obtain accepted tokens (i.e., ``auto'' and ``regressive''), and $k$ new candidate tokens (i.e., ``model'' and ``<s>'') are generated from one of the mask groups after a single model invocation. An illustration of the generation and verification process can be found in Figure~\ref{fig:llm_forward}. SPACE allows for a variable number of tokens to be generated in each step, with the quantity ranging from a minimum of 1 to a maximum of $k+1$. }
\label{fig:space_process}
\end{figure*}

In this paper, we propose Smart Parallel Auto-Correct dEcoding (SPACE), a approach that allows LLMs to generate multiple tokens speculatively while simultaneously verifying them. SPACE harmonizes a SAR model with a draft-then-verify inference algorithm to optimize inference speed while maintaining high model quality. We demonstrate that an AR language model can be adapted to produce probable token sequences in parallel through semi-autoregressive supervised fine-tuning (SAR-SFT). This strategy obviates the need for supplementary  models and maintains the fine-tuning process within reasonable computational demands. 
We also introduce an auto-correct decoding algorithm that enables the generation and validation of token candidates to occur concurrently within a single invocation of a model, thereby significantly boosting inferential efficiency. SPACE is particularly useful for edge server applications of LLMs, where it can effectively utilize the computing resources to accelerate the inference speed in low batch size scenarios.
A visual comparison between AR and SPACE inference can be found in Figure~\ref{fig:space_process}.
%To accelerate the inference speed, we employ the draft-then-verify method and introduce auto-correct decoding algorithm. This innovative algorithm enables the generation and validation of token candidates to occur concurrently within a single invocation of the model, thereby significantly boosting inferential efficiency.
%This pioneering algorithm facilitates the concurrent generation and verification of candidate tokens within a single model invocation, significantly enhancing the model's inferential efficiency. 
Our key contributions are summarized as follows:
\begin{itemize}
\item  We propose a SAR-SFT scheme that empowers autoregressive LLMs to generate multiple  tokens at once, without requiring substantial  computational overhead.
  
\item  We introduce an auto-correct decoding algorithm that facilitates the concurrent generation and validation of candidate tokens within a single forward pass of the model.
%, ensuring a lossless acceleration in inference speed. %Our approach is distinguished by its memory efficiency and user-friendly, as it obviates the need for ancillary models or additional trainable parameters.

\item  Our extensive experiments, conducted across various LLMs with parameters ranging from 6B to 70B, validate that SPACE is effective in achieving an inference speedup from 2.7x to 4.0x in HumanEval-X while maintaining output quality. 

\end{itemize}
%This makes SPACE a competitive method for enhancing inference efficiency in LLMs.
%This speedup is on par with, if not surpassing,  alternative approaches such as speculative decoding, demonstrating the efficacy of SPACE as a competitive technique for enhancing inference efficiency in LLMs.

%The remainder of this paper is structured as follows: Section 2 reviews related work in the field. Section 3 details the SPACE methodology. Section 4 presents our experimental findings. Section 5 offers concluding remarks, and Section 6 discusses the limitations of SPACE.

\section{Related Work}
% LLMs play a critical role in present-day AI applications, making the optimization of LLM inference a crucial area of focus.  A wealth of systems-level optimizations have been proposed to increase the throughput of LLM services. Our paper specifically delves into a crucial line of research involving speculative decoding and semi-autoregressive decoding, two approaches that have shown promise in enhancing inference efficiency.
%LLMs play a critical role in present-day AI applications, making the optimization of LLM inference a crucial area of focus. 
%Numerous optimizations have been proposed to enhance the inference speed of LLMs. 
% Our paper explores a vital line of research involving speculative decoding and semi-autoregressive decoding—two approaches that have demonstrated promise in improving inference efficiency.

\textbf{Speculative Decoding}
Speculative decoding~\cite{google2023fast, deepmind2023accelerating} accelerates LLM inference by using a smaller draft model to predict larger target model outputs, with subsequent verification by the target model. The efficacy of the method is contingent on the accuracy of the draft model's predictions. To enhance accuracy, researchers have adopted various strategies such as employing ensembles of boosted draft models~\cite{miao2023specinfer}, staged draft models~\cite{spector2023accelerating}, retraining the target model with addition of auxiliary prediction heads~\cite{stern2018blockwise}, introducing advanced coordination policies~\cite{kim2023speculative} and refining the decoding algorithm~\cite{sun2023spectr, lin2024bita}. 
%These approaches generally presume a fixed draft model after deployment. Conversely, some advocate for dynamically adjusting the draft model in response to evolving user queries during runtime, thereby making efficient use of surplus computational resources in the serving cluster~\cite{liu2023online}.
However, speculative decoding hinges on the accessibility of suitable smaller models, which can be difficult to obtain and often requiring extra training and careful tuning~\cite{liu2023online}. SPACE circumvents this challenge by fine-tuning the target model to prognosticate future token sequences in parallel, eliminating the dependency on extra small model. %Despite this, SPACE manages to achieve speedups comparable to, if not surpassing, that of speculative decoding while preserving output quality.

Recent advancements like Lookahead Decoding~\cite{fu2023lookahead} and Self-Speculative~\cite{zhang2023draft} have refined the draft-then-verify process, forgoing the need for extra models or intricate training steps. Although simpler, these methods tend to provide less acceleration than SPACE. Contrarily, Medusa~\cite{cai2024medusa} and PaSS~\cite{monea2023pass} leverage fine-tuning of a single LLM to perform both token generation and validation. PaSS, through fine-tuning lookahead token embeddings, allows LLMs to predict multiple tokens ahead, but with a limited speedup of about 30\% as their verification and drafting phases occur sequentially. Medusa adds multiple decoding heads to the LLM and introduces a tree-based decoding algorithm for faster inference, but it struggles with larger batch sizes owing to the increased computational demands of tree-structured attention.
%are appealing as they simplify the draft-then-verify process by not necessitating auxiliary models or additional training regimes. However, these methods typically achieve a lower speedup ratio when compared to SPACE. Methods such as Medusa~\cite{cai2024medusa} and PaSS~\cite{monea2023pass} suggest innovative approaches where a single LLM, with certain fintuning, is able to generate and verify tokens. PaSS~\cite{monea2023pass} finetunes the lookahead token embedding to enable LLM to generate multiple future tokens. However, PaSS achieves much lower speedup (around 30\%) as their verification and drafting phases occur sequentially. Medusa~\cite{cai2024medusa} introduces multiple decoding heads on top of the LLM backbone, and they propose a tree-base decoding algorithm to speedup inference. Their method, while effective with small batch size, can not perform well in large batch size scenario as the tree attention introduces much more computational demand. 

\textbf{Semi-Autoregressive Decoding}
SAR departs from the AR approach by decoding multiple tokens in parallel, thereby significantly enhancing inference efficiency. Particularly in machine translation, SAR has achieved a fivefold speed increase while preserving 88\% of the model quality~\cite{wang2018semi}. 
%Recent research efforts to enhance SAR performance in machine translation include employing alignment-focused training objectives~\cite{gu2020fully}, innovating model architectures~\cite{huang2022directed}, etc. However, to the best of our knowledge,  exploration of SAR in conjunction with decoder-only LLMs remains limited.
For SAR decoding, it is a common trick to employ mask tokens as placeholders in input. 
This approach, originating from the mask-predict paradigm introduced by \citeauthor{ghazvininejad2019mask} in machine translation, has since become a widely recognized decoding strategy \cite{xiao2023survey}. 
%In fact, the mask-predict paradigm was initially introduced by~\citeauthor{ghazvininejad2019mask} in  machine translation and has since gained widespread acceptance as a decoding strategy~\cite{xiao2023survey}. 
Inspired by this paradigm, SPACE adopts $k$ mask tokens to predict $k$ future tokens. \citeauthor{xia2022lossless} accelerate inference of an AR model by creating draft tokens with a SAR model and then refined by the AR model. Unlike their method which necessitates a secondary SAR model for draft generation, SPACE eliminates this requirement, merging generation and verification phases for enhanced efficiency. 
%In addition, SpecDec~\cite{xia2022lossless} adopts more flexible criteria for draft token validation by permitting the top-k tokens to pass through. SPACE, on the other hand, maintains rigorous validation standards as in speculative decoding~\cite{deepmind2023accelerating, google2023fast}.
%implements more lenient criteria for validating draft tokens, an angle that could mesh well with tasks such as machine translation but might hold back applications where the threshold for errors is lower, including certain chatbots or content creation tools. Our SPACE does not introduce such relaxation.
%More discussions on the comparisons between SpecDec~\cite{xia2022lossless} and SPACE can be found in Appendix~\ref{app:previous_work}.

%One worth noting work that use SAR model to accelerate inference is SpecDec~\cite{xia2022lossless}, which employs a strategy of initially decoding a block of tokens quickly as a draft with a SAR model before refining this draft using an AR model. However, a notable difference is that SpecDec requires a secondary draft SAR model to produce candidate tokens, but SPACE obviates this requirement. Furthermore, SPACE is distinguished by its simultaneous generation and verification process, a distinct improvement over ~\citeauthor{xia2022lossless}’s method.
%However, a notable difference is that SpecDec requires the training of an extra SAR model, which introduces resource overhead.

\section{Methods}
SPACE primarily comprises two components: the SAR-SFT scheme and the auto-correct decoding algorithm. 
The SAR-SFT scheme enhances an autoregressive LLM's capacity for speculative multi-token generation in a single decoding step. Meanwhile, the auto-correct decoding algorithm allows the LLM to concurrently generate and verify candidate tokens. 
% This integrated approach ensures equivalence in the ultimate token distribution output, effectively reproducing the results that would be achieved through the standard AR inference process of the model. 
We introduce the details of these two components in the following subsections.
\begin{figure*}[!th]
\centering
\includegraphics[width=5.8in]{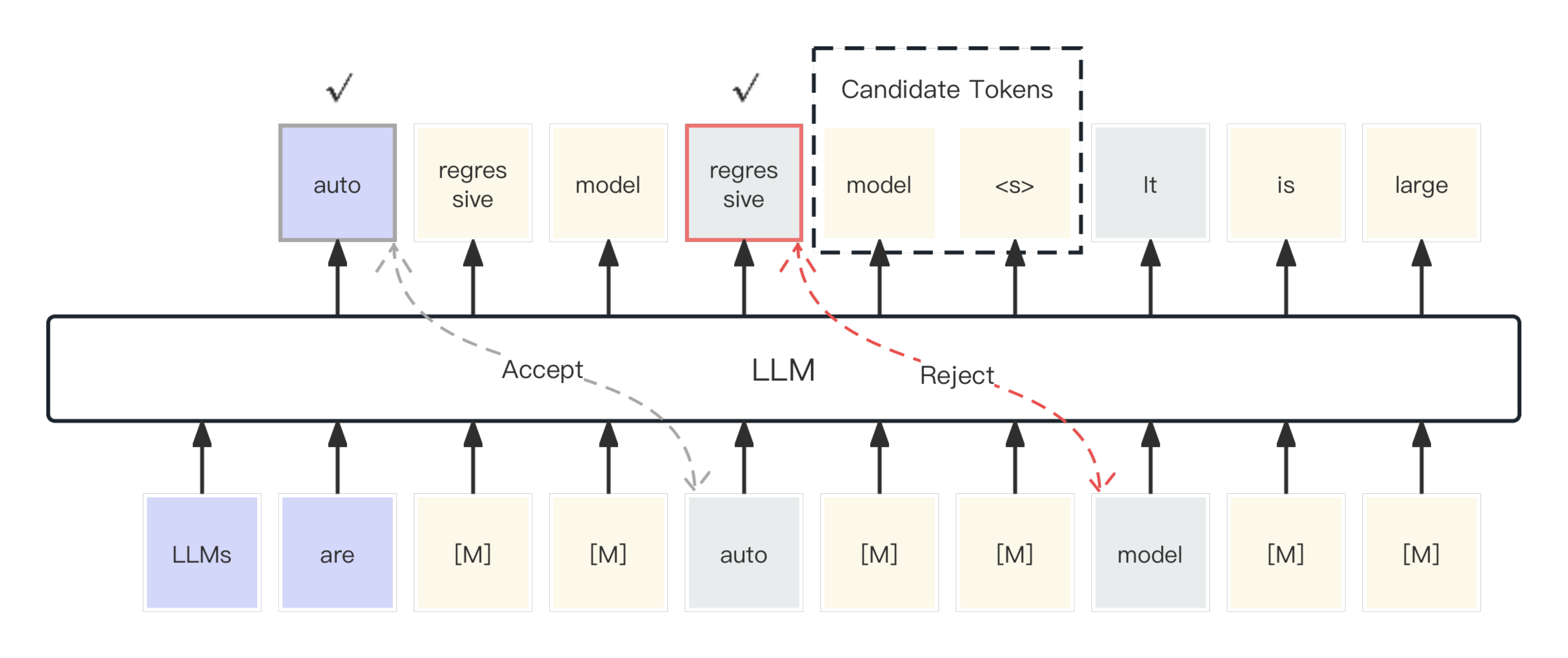}
\caption{An illustrative example of the auto-correct decoding algorithm in SPACE. In this example, the first candidate token ``auto'' is accepted, while the second candidate token ``model'' is rejected. The LLM generates two new tokens ``auto'' and ``regressive'' in this decoding step and two new candidate tokens ``model'' and ``<s>'' from the second mask group.}
\label{fig:llm_forward}
\end{figure*}

\subsection{Semi-Autoregressive Finetuning}
Conventionally a pretrained LLM undergoes a process known as supervised fine-tuning (SFT) to adapt the model to specific downstream tasks. Specifically, given the prompt token sequence $X$ and the answer token sequence $Y=\{y_1, y_2, \cdots, y_N\}$, the AR model is trained in SFT with loss function
\begin{equation}
\mathcal{L}_{AR}=-\sum_{t=1}^N\log P(y_t|y_{<t}, X; \theta),
\end{equation}
where $y_t$ is the token to be predicted at step $t$, $y_{<t}$ is the tokens predicted in previous $t-1$ decoding steps and $\theta$ is the model parameters.

% In our proposed semi-autoregressive training, we randomly sample an index $i^*$ from $\{1, 2, \cdots, N\}$ and append $k$ consecutive target tokens starting from $i^*$ to $Y$, meanwhile we appends $k$ special ``[M]'' tokens to $X$. The source and target token sequences now become:
% \begin{equation}
% \begin{aligned}
%     % X &= \{x_1, x_2, \cdots, x_M, \underbrace{[M], \cdots, [M]}_{\times k}\} \\
%     Y &= \{y_1, y_2, ..., y_N, y_{i^*}, \cdots, y_{i^*+k}\}
% \end{aligned}
% \end{equation}
% and train the model to predict these masked tokens in parallel:
In the proposed SAR-SFT scheme, our objective is to train the model to generate $k$ consecutive tokens when presented with an input sequence containing $k$ mask tokens. The adaptation from traditional SFT to SAR-SFT affects only the dataloader component in implementation. In this modified dataloader, each data sample remains unchanged with a probability $p_{\mathrm{ar}}$. Conversely, with a probability of $1-p_{\mathrm{ar}}$, we randomly select a position $m$ from $\{0, 1, \cdots, N-k\}$ in the input sequence to replace $k$ consecutive tokens with mask tokens. We then truncate the input token sequences to keep the first $m+k$ tokens, denoted as $y_{<m}^k$:
%To achieve this, we employ an autoregressive loss $\mathcal{L}_{AR}$ with a probability $p_{\mathrm{ar}}$. Conversely, with a complementary probability of $1-p_{\mathrm{ar}}$, we randomly sample an index $m$ from $\{0, 1, \cdots, N-k\}$ and obtain $y_{<m}$ from the answer token sequence. Subsequently, we append $k$ mask tokens ``[M]''  to $y_{<m}$ to form $y_{<m}^k$:
\begin{equation}
y_{<m}^k = \{y_1, y_2, \cdots, y_{m-1}, \underbrace{[M], \cdots, [M]}_{\times k}\}.  
\end{equation}
Under this modified dataloader, with probability $p_{\mathrm{ar}}$ the model is trained with the original AR loss $\mathcal{L}_{AR}$. With probability $1-p_{\mathrm{ar}}$, the model is trained with the SAR loss function defined as follows:
\begin{align}
    \mathcal{L}_{SAR} &= - \sum_{t=1}^{m-1}\log P(y_t|y_{<t}, X; \theta) \nonumber \\&\quad  - \sum_{t=m}^{m+k}\log P(y_t|y_{<m}^k, X; \theta) 
\end{align}
% The SAR loss is designed to 
%The final loss function we used in SAR-SFT is
%\begin{equation}
%    \mathcal{L} = p_{\mathrm{ar}} \mathcal{L}_{AR}+(1-p_{\mathrm{ar}}) \mathcal{L}_{SAR}
%\end{equation}
Intuitively, the hyper-parameter $p_{\mathrm{ar}}$ plays a critical role in striking a balance between the AR loss and the SAR loss. By selecting an appropriate value for $p_{\mathrm{ar}}$, the LLM is trained not only to adhere to downstream tasks but also to predict multiple tokens at each decoding step.
%With a proper $p_{\mathrm{ar}}$, the LLM learns to not only follow the instructions but also predict multiple tokens at each decoding step. 

We note that the primary goal of SAR-SFT is not to compel the LLM to predict several tokens in parallel with high accuracy, as this can be an exceedingly challenging task. Rather, our goal is to enable the LLM to make an ``educated guess'' about the upcoming few tokens, which is more attainable. %This strategy not only reduces training expenses but also enhances the model's predictive efficiency.
%as achieving such a task can be exceedingly challenging and requiring much more computation resource to train the model. Instead, to reduce training cost, we aim for the LLM to make an ``educated guess'' about the upcoming few tokens, which is more attainable. 
%and aligns with the idea of equipping the model with the ability to speculatively generate predictions based on its current understanding of the context, rather than predicting with exact precision. 
%We note that the objective of SAR-SFT is to compel the LLM to accurately predict several tokens in parallel. However, accomplishing this task with high accuracy presents significant difficulty. Instead, it is more attainable for the LLM to make an ``educated guess'' about the upcoming few tokens. 
This strategy allows the model to improve its inference efficiency by preparing probable token sequences beforehand, which can later be validated and refined by the auto-correct decoding algorithm introduced in next subsection. %An illustrative example is shown in Figure~\ref{fig:llm_forward}.

\subsection{Auto-Correct Decoding Algorithm}
%Typically a SAR model possesses the capability to facilitate inference in both AR and SAR manners. However, when deploying SAR inference, there is usually a noticeable degradation in the model's performance~\cite{xiao2023survey}. Conversely, utilizing AR inference results in commendable performance, albeit at a diminished speed.
%Inspired by the idea of speculative execution to accelerate AR decoding~\cite{deepmind2023accelerating, google2023fast, stern2018blockwise}, we propose an efficient draft-then-verify inference algorithm to speedup the AR inference of the SAR model.

% Different from previous methods~\cite{google2023fast, deepmind2023accelerating, miao2023specinfer, xia2022lossless} that depended on auxiliary models to produce candidate tokens, SPACE employs the same model for both the generation and verification of candidate tokens. To improve inference efficiency, we design an inference algorithm that empowers the same LLM to verify candidate tokens from the preceding step and produce new candidates for the subsequent step within a singular forward pass. Since LLM decoding is primarily bounded by memory bandwidth, we can merge the token generation and verification in the same decoding step, leveraging GPU's parallel processing power to hide overheads. 
Unlike previous methods~\cite{google2023fast, deepmind2023accelerating} that rely on auxiliary models, SPACE streamlines the process by using the same LLM for generation and subsequent verification of candidate tokens. To enhance inference efficiency, we have developed an algorithm that enables this unified LLM to concurrently verify tokens from the current step and generate new candidates for the next step within a single forward pass. 
%Given that LLM decoding is primarily bounded by memory bandwidth, our method effectively harness the GPU's parallel processing capabilities to minimize latencies and maximize computational efficiency.
%enable the same LLM to verify candidate tokens from previous step and generate new candidate tokens for next step in single forward pass. 

\begin{algorithm}
\caption{The auto-correct decoding algorithm}
\label{alg:infer}
\begin{algorithmic}[1]
\Require A sequence of input tokens $\mathcal{T}$, number of mask tokens $k$, large language model $\mathcal{M}$
\Ensure A sequence of  generated tokens $\mathcal{O}$
\State $\mathcal{O}=\mathcal{T}$, $L_c=[0]\times k$, $P_c=[+\infty]\times k$
\While{True}
\State $l=len(\mathcal{O})$
% \State $\mathcal{I} = \mathcal{I} + [M]\times k$ \Comment{Append input with list of k mask tokens}
% \For{$i=1$ to $k$} %\Comment{Construct new input token ids}
% \State $\mathcal{I}=\mathcal{I}+  L_c[i]+[M]\times k $
% \EndFor
%\State $\mathcal{I}, \bar{A}, \bar{P} = GetInput(l, k)$ \Comment{Get extended input, attention mask and positional encoding according to equation~(\ref{eq:extend_input})-(\ref{eq:position})}
\State Get $\mathcal{I}, \bar{A}, \bar{P}$ according to equation~(\ref{eq:extend_input})-(\ref{eq:attention_mask})
\State $P=\mathcal{M}(\mathcal{I}, \bar{A}, \bar{P})$ \Comment{Get the output logits}
\State $idx = l+1$
\State $Q = P[l]$  \Comment{The logit of the $l$-th token}
\For{$i=1$ to $k$}
\State $r\sim U(0, 1)$
\If{$r\leq Q(L_c[i])/P_c[i]$} %\Comment{Accept candidate token $L_c[i]$}
\State $\mathcal{O}.append(L_c[i])$
%\If{<EOS> in $\mathcal{O}$}
%\State return $\mathcal{O}$
%\EndIf
\State $idx=idx+k+1$
\State $Q=P[ l + i * (k+1)]$
\Else 
\State break
\EndIf
\EndFor
\State $a\sim Q$ \Comment{Sample one extra token}
\State $\mathcal{O}.append(a)$
\If{<EOS> in $\mathcal{O}$}
\State return $\mathcal{O}$[:eos\_index]
\EndIf
%\State \Comment{Get new tokens and their probability}
\State $L_c \sim P[idx: idx+k]$  \Comment{New candidates}
\State $P_c = P[idx: idx+k](L_c)$ \Comment{Probability} 
\EndWhile
\end{algorithmic}
\end{algorithm}

Algorithm~\ref{alg:infer} outlines the auto-correct decoding algorithm employed in SPACE, with Figure~\ref{fig:llm_forward} providing an illustrative example. When presented with an initial prompt (e.g., ``LLMs are'') alongside $k$ candidate tokens (e.g., ``auto'' and ``model''), the algorithm begins by constructing an input token sequence. This is achieved by augmenting the original prompt with $k+1$ groups of mask tokens, interspersed with the $k$ candidate tokens, as depicted in Figure~\ref{fig:llm_forward}. Specialized attention masks and positional indices are devised to constrain the influence of the mask tokens, allowing them to ``see'' only preceding non-mask tokens and other mask tokens within their respective groups. Following a forward pass through the LLM, a verification step is applied to the candidate tokens. If $i^*$ tokens (where $0 \leq i^* \leq k$) pass this check, the algorithm proceeds to generate $k$ new candidate tokens (e.g., ``model'' and ``<s>'') from the ($i^*+1$)-th mask group and one extra token from the $i^*$-th candidate token (e.g., ``regressive''). In this case, $i^*+1$ new tokens are generated from a single LLM forward step.

%Algorithm~\ref{alg:infer} outlines the auto-correct decoding algorithm used in SPACE and Figure~\ref{fig:llm_forward} gives an illustrative example. In general, when given an initial prompt (e.g., ``LLMs are'') along with $k$ candidate tokens (e.g., ``auto'' and ``model''), the algorithm begins by forming an input token sequence. This is achieved by augmenting the original prompt with $k+1$ groups of mask tokens, interspersed with the $k$ candidate tokens, as depicted in Figure~\ref{fig:llm_forward}. We devise specialized attention masks and positional indices to confine the scope of the mask tokens, enabling them to see'' only preceding non-mask tokens and other mask tokens within their respective groups. Following a forward pass through the LLM, a verification step is applied to the candidate tokens. If $k^*$ tokens (where $0 \leq k^* \leq k$) pass this check, we then proceed to generate $k$ new candidate tokens (e.g., ``model'' and ``<s>'') from the ($k^*+1$)-th mask group and one extra token from the $k^*$-th candidate token (e.g. ``re). In this case, $k^*+1$ new tokens is generated from a single LLM forward step.

%We note that this decoding algorithm is applicable to both greedy and random sampling settings. Since greedy sampling can be considered a special case of random sampling, we introduce the auto-correct decoding algorithm within the broader context of random sampling setting without loss of generality.

In the following, we introduce the auto-correct decoding algorithm in details. Given a sequence of input prompt tokens $\mathcal{T}=\{x_1, x_2, \cdots, x_l\}$ and a list of $k$ candidate tokens $L_c=\{c_1, c_2, \cdots, c_k\}$ generated from the previous decoding step, we first construct a sequence of input tokens $\mathcal{I}$ as follows:
% \begin{equation}
% \begin{aligned}
% \label{eq:extend_input}
%     \mathcal{I} = & \{x_1, x_2, \cdots, x_l,\\& L_m^k, c_1, L_m^k, c_2, \cdots, c_k, L_m^k\},
% \end{aligned}
% \end{equation}
\begin{equation}
\label{eq:extend_input}
    \mathcal{I} =  \{x_1, \cdots, x_l, L_m^k, c_1, L_m^k, \cdots, c_k, L_m^k\},
\end{equation}
where $L_m^k=\underbrace{[M],\cdots,[M]}_{\times k}$ represents a group of $k$ mask tokens and there are $k+1$ groups of them in $\mathcal{I}$. The sequence $\mathcal{T}$ is expanded by $k \cdot (k+2)$ additional tokens, resulting in a total length of $|\mathcal{I}| = l + k \cdot (k+2)$. These $k+1$ groups of mask tokens are designated for the generation of new candidate tokens. %Depending on the number of accepted tokens, the predictions from one of the mask token groups is chosen as new candidate tokens.

%The input token sequence is extended by $k*(k+2)$ additional tokens and its total length is $|\mathcal{I}|=l+k*(k+2)$. These $k+1$ groups of mask tokens are responsible for new candidate tokens generation. Depending on the number of accepted tokens, we will choose the generation results from one of mask tokens groups as candidate tokens, the details will be shown later.
%Typically, $l \gg k$ and thus the additional computational cost is neglectable.

Since LLM decoding is primarily bounded by memory bandwidth, we can merge the generation and verification in the same forward step, leveraging GPU's parallel processing power to hide overheads. We achieve this by designing special attention mask $\bar{A}\in\{0, 1\}^{|\mathcal{I}|\times |\mathcal{I}|}$  as follow:
\begin{equation}
\label{eq:attention_mask}
    \bar{A}_{ij} = \left\{
\begin{aligned}
1 & \quad i\geq j, \mathcal{I}[j]\neq M \\
1 & \quad i\geq j, i-j< k, \mathcal{I}[i]=\mathcal{I}[j]= M \\
0 & \quad otherwise
\end{aligned}
\right.
\end{equation}
% \begin{equation}
% \label{eq:position}
%     \bar{P}_i = \sum_{j=1}^{|\mathcal{I}|}A_{ij}-1
% \end{equation}
%The attention mask is tailored such that mask tokens are limited to causally attend only to their counterparts within the same group and to the preceding non-masked tokens. In addition, all the non-masked tokens only have causal attentions on previous non-masked tokens and they can not attend to previous mask tokens.
% The attention mask is tailored such that masked tokens can causally attend only to other mask tokens within the same group and to preceding non-masked tokens. Furthermore, all non-masked tokens are restricted to causally attend to prior non-masked tokens, and are unable to attend to any preceding masked tokens.
%This ensures that the model's generation process remains structured and preserves the causal sequence of the tokens. 
An illustrative example of the attention mask configuration is depicted in Figure~\ref{fig:llm_attention} in Appendix~\ref{app:attention}. The positional indices for positional encoding can be computed as $\bar{P}_i = \sum_{j=1}^{|\mathcal{I}|}\bar{A}_{ij}-1$.

%The extended input token sequence $\mathcal{I}$, together with attention mask $\bar{A}$ and positional encoding $\bar{P}$, are passed through the LLM. This facilitates the inference process, allowing the LLM to generate the normalized output logits, denoted as $P$, as outlined in line 5 of Algorithm~\ref{alg:infer}.

Following the input construction phase, we proceed with the inference process using the LLM, from which we derive the normalized output logits, denoted as $P$.
The candidate tokens are then verified through rejection sampling, which is detailed from line 6 to line 22 in Algorithm~\ref{alg:infer}. 
Denote $P_c$ as the list of semi-autoregressive probability of candidate tokens obtained from the previous step. Formally $P_c[i]$ is defined as:
\begin{equation}
    P_c[i] = P(c_i|x_1, \cdots, x_{l-1}, \underbrace{[M], \cdots, [M]}_{\times i})
\end{equation}
Denote $Q_c$ as the list of autoregressive probability of candidate tokens from the current step~\footnote{By definition, $Q_c[i]$ is equivalent to $Q(L_c[i])$ in line 10 of Algorithm~\ref{alg:infer}.}. 
\begin{equation}
    Q_c[i] = P(c_i|x_1, \cdots, x_{l}, c_1, \cdots, c_{i-1})
\end{equation}
Starting from $i=1$, we accept token $c_i$ with probability $\min(1, \frac{Q_c[i]}{P_c[i]})$.
%This can be implemented by first sample a random number uniformly from $[0,1]$, and then accept the token if this random number does not exceed the ratio $Q_c[i]/P_c[i]$. 
Upon acceptance of token $c_i$, the algorithm output $c_i$ and proceeds to validate the subsequent token $c_{i+1}$ using the same criterion; conversely, if $c_i$ is rejected, the verification process terminates immediately and one extra token is generated from the output logit of the last accepted candidate token, as shown in line 18 in Algorithm~\ref{alg:infer}. It is important to observe that during each decoding step, the number of generated tokens ranges from a minimum of one to a maximum of $k+1$. By employing rejection sampling, it can be proved that the distribution of the output token sequence matches that of the AR inference process in the LLM. For a more comprehensive explanation of this claim, readers can refer to prior research~\cite{google2023fast, deepmind2023accelerating}.

\section{Experiments}
\subsection{Experimental Settings}
% \subsubsection{Training} 
\textbf{Training}
We conduct experiments on LLMs with various sizes, including ChatGLM3-6B-Base~\cite{du2022glm}, LLaMA-2 (7B, 13B, 70B)~\cite{touvron2023llama}, Qwen-14B~\cite{qwen}, InternLM-20B~\cite{2023internlm}, Falcon-40B~\cite{almazrouei2023falcon}. To ensure reproducibility, we finetune the models using publicly available SFT datasets, including Alpaca-GPT4~\cite{peng2023instruction}, Lima~\cite{zhou2023lima}, Oaast-SFT~\cite{OpenAssistant}, CodeAlpaca~\cite{codealpaca}, and OpenPlatypus~\cite{platypus2023}. 
The details of these datasets are listed in Table~\ref{tab:dataset} in Appendix~\ref{app:table}. There are in total 166,993 training samples. 
We add the mask token as a special token and initialize its embedding with normal distribution. Unless otherwise specified, we set the number of mask tokens $k=5$ and $p_{\mathrm{ar}}=0.5$. %We finetune the models for 2 epochs with a learning rate as 5e-5. 
The training details can be found in Appendix~\ref{app:table}.

% \subsubsection{Inference} 
\textbf{Inference}
In our assessment of SPACE, we employ four distinct datasets: Chatbot Instruction Prompt (CIP) \cite{cip2023}, MT-Bench \cite{zheng2023judging}, HumanEval-X \cite{zheng2023codegeex} and XSum \cite{narayan2018don}. 
%CIP is a conversational dataset from which we utilize prompts to simulate realistic conversations. MT-Bench is a dataset comprised of multi-turn questions, encompassing a wide range of topics. HumanEval-X is a standard benchmark for Python code generation and Pass@10 is used as the metric. Lastly, the XSum dataset, which tasks models with summary generation, is evaluated using  ROUGE-L. 
%This metric measures the longest common subsequence between the generated summary and a reference summary, providing an indicator of the overlap in terms of sentence-level structure and content fidelity. 
%We use the generation algorithm from the Huggingface Transformers library~\cite{wolf-etal-2020-transformers} as the baseline method and run inference in an auto-regressive manner. To measure the inference efficiency of SPACE, we employ two metrics: speedup and average accepted tokens. The speedup metric quantifies computational efficiency, calculated as the ratio of the baseline generation algorithm's inference time to the inference time achieved when using SPACE. On the other hand, average accepted tokens serves as an indicator of generation efficacy. It is defined as the ratio of the total number of tokens produced to the number of inference steps that the large language model (LLM) executed. This metric provides insights into how productively SPACE utilizes each LLM inference step for token generation. 
For inference baseline, we train LLMs with SFT under the same datasets and training configuration used for SAR-SFT. We adopt the generation algorithm provided by the Huggingface Transformers library~\cite{wolf-etal-2020-transformers}, executing it in an autoregressive fashion on the SFT model. We conduct the experiments on a server with eight A800 (80GB) GPUs. By default, we set the batch size to 1 during inference. To evaluate the inference efficiency of SPACE, we employ two metrics: speedup and average accepted tokens. The speedup metric is defined as the ratio of the inference speed of the baseline method (measured in tokens per second) to the inference speed achieved using SPACE. %Essentially, this metric quantifies the relative increase in speed that SPACE provides over the baseline approach. 
The second metric, average accepted tokens, is computed as the ratio of the total number of tokens generated to the number of inference steps performed by the LLM. %This provides a measurement of how effectively SPACE makes use of each inference step, concretely demonstrating the method's ability to produce a greater number of tokens per step and thereby contribute to an overall increase in efficiency. 
The evalutation details can be found in Appendix~\ref{app:eval}.
%CIP is a conversational dataset and we use the prompts from the dataset to simulate the real-world conversation trace. MT-Bench contains a suite of challenging multi-turn questions that are open-ended by nature, covering various topics including coding, extraction, humanities, math, reasoning, roleplay, stem and writing. HumanEval-X is a code generation task and we evaluate the model on python code completion tasks using metric Pass@1. Lastly, XSum is a natural language summarisation task and it is evaluated using ROUGE-L. %Our evaluation covers these diverse datasets to ensure a comprehensive analysis of SPACE's capabilities.

\subsection{Experimental Results}
\subsubsection{Inference Efficiency} 

\begin{table*}[htp]
\centering
\begin{tabular}{r | c | c | c | c | c | c | c | c}
\hline
\multirow{2}{*}{\textbf{Model}} & \multicolumn{3}{c |}{\textbf{XSum}} & \multicolumn{3}{c|}{\textbf{HumanEval-X}} & \multicolumn{2}{c}{\textbf{CIP}} \\
\cline{2-9}
& ROUGE-L & \makecell[c]{Avg. \\ Tokens}  &  \makecell[c]{Speed-\\ up} & Pass@10 & \makecell[c]{Avg.\\ Tokens}  &  \makecell[c]{Speed-\\ up} & \makecell[c]{Avg. \\ Tokens}  &  \makecell[c]{Speed-\\ up}\\
\hline
ChatGLM-3-6B  & 14.5 (14.3) & 2.04 &1.48 &18.3 (18.3)& 3.34 & 2.71 & 1.80 & 1.52 \\
LLaMA-2-7B &  16.0 (16.1) & 2.23 & 1.92 & 18.9 (18.9)& 3.54 & 3.18 &1.85 & 1.72\\
LLaMA-2-13B&15.1 (15.0) & 2.36 & 2.08 & 20.1 (20.1) & 3.76 & 3.43 & 1.99 & 1.82\\
Qwen-14B & 17.2 (17.2) & 2.15 & 1.94 & 26.8 (26.8) & 3.51 & 3.19 & 1.85 & 1.68 \\
InternLM-20B & 16.4 (16.3) & 2.15 & 1.99 & 21.3 (21.3) & 3.31 & 3.16 & 1.80 & 1.63 \\
Falcon-40B & 15.7 (15.8) & 2.17 & 2.03 & 20.7 (20.7) & 3.58 & 3.58 & 1.96 & 2.02 \\
LLaMA-2-70B & 16.4 (16.3) & 2.54 & 2.34 & 28.0 (28.0) & 4.32 & 4.04 & 2.09 & 1.89 \\
\hline
\end{tabular}
\caption{The experimental results on XSum, HumanEval-X and CIP under greedy sampling setting. We show the average accepted tokens (Avg. Tokens) and inference speedup (Speedup) for each datasets. The number in parentheses shows the corresponding results of the baseline method.}
\label{tab:result_xsum_humaneval_cip}
\end{table*}

The experimental results on XSum, HumanEval-X and CIP under greedy sampling setting are shown in Table~\ref{tab:result_xsum_humaneval_cip}. 
%Under greedy decoding conditions, we anticipate identical outputs from models applying SPACE and baseline autoregressive generation method. However, the results exhibit occasional discrepancies between the two, which can be attributed to numerical variations during decoding that cause the generation of different tokens, potentially leading to significant divergence in the resulting sequences. Despite these observed differences, 
We observe that SPACE predominantly corresponds closely with baseline performance levels in both the XSum and HumanEval-X benchmarks. Moreover, SPACE demonstrably realizes a speedup in the range of 1.5 to 4.0, depending on the models and datasets. The maximal acceleration, seen in LLaMA-2-70B on HumanEval-X, clocks in at an impressive 4.04. More experimental results of SPACE under random sampling can be found in Appendix~\ref{app:random_sampling}. 

From the above results, we have the following three observations: First, SPACE consistently achieves speedup while maintaining performance comparable to the baseline across models of varying sizes, showcasing its broad applicability. Specifically, the results attained by SPACE in tasks such as XSum and HumanEval-X closely mirror those achieved by the baseline method, as indicated by the comparable performance metrics listed in parentheses in Table~\ref{tab:result_xsum_humaneval_cip}.

Second, the magnitude of speedup experienced is model-specific, indicating that the efficiency benefits of SPACE can differ in models. 
%This could be due to two factors: firstly, models have different vocabularies, with less efficient vocabularies potentially leading to higher speedup as they render tokens more predictable. Secondly, models with more parameters tend to show greater speedup, possibly because their enhanced ability permits better anticipation of upcoming tokens.  
This variance might stem from several factors: (1) the models' vocabularies vary, with less efficient vocabularies possibly leading to greater predictability and thus higher speedup; and (2) models with more parameters often enjoy more substantial speedup, likely owing to their superior predictive capabilities that facilitate earlier anticipation of forthcoming tokens.

Lastly, when applying SPACE to different tasks, the same model can exhibit dramatically different speedup ratios. In particular, tasks that involve programming, such as those in the HumanEval-X benchmark, exhibit the most significant speedup, achieving an average rate of 3.33 using greedy sampling. This observation aligns with the results in previous research~\cite{deepmind2023accelerating}, and could be attributed to the inherently structured and predictable nature of programming code.

%Second, a singular model exhibits markedly varied speedup ratios when applied to diverse tasks. Remarkably, tasks involving coding, such as those in the HumanEval-X benchmark, register the highest speedup, boasting an average rate of 3.36 when using greedy sampling. This observation aligns with the results in previous research~\cite{deepmind2023accelerating}, and could be attributed to the inherently structured and predictable nature of programming code.

%The speedup varies across different models, with a general trend indicating that models with a larger number of parameters exhibit greater speedup. This could be attributed to the increased capability of larger models to anticipate subsequent tokens more effectively.

%We conduct ablation study on the number of mask token $k$ on LLaMA-2-7B. The results are depicted in Figure~\ref{fig:ablation_k}. We observe that the best number of mask tokens is 5.  During the SAR-SFT stage, the LLM is trained to simultaneously predict consecutive $k$ tokens. As $k$ increases, the prediction task becomes more challenging, leading to a decrease in speedup. Conversely, if $k$ is too small, the benefits of parallel decoding become less significant.

To ensure a fair and unbiased comparison, we have reproduced several accelerating methods, such as self-speculative decoding~\cite{zhang2023draft}, look-ahead decoding~\cite{fu2023lookahead}, assistant generation from HF~\cite{gante2023assisted} and speculative decoding~\cite{google2023fast}. All experiments were executed on the MT-Bench dataset using identical hardware configurations in conjunction with the LLaMA-2-70B model. According to the outcomes presented in Table~\ref{tab:result_baseline}, SPACE outperforms other methods by achieving the highest speedup. This demonstrates the competitiveness of SPACE in inference acceleration.
\begin{table}[htp]
\centering
\begin{tabular}{l | c  }
\hline
\textbf{Method} &\textbf{Speedup}  \\
\hline
Self-SpecDec~\cite{zhang2023draft} & 1.15 \\
Look-ahead~\cite{fu2023lookahead} & 1.22 \\
HF AssistGen~\cite{gante2023assisted} & 1.53 \\
SpecDec~\cite{google2023fast} & 1.79 \\
SPACE (ours) & 2.26 \\
\hline
\end{tabular}
\caption{Comparison of speedup for various acceleration methods with LLaMA-2-70B on MT-Bench dataset.}
\label{tab:result_baseline}
\end{table}

\subsubsection{Impact of SAR-SFT on Model Quality}
While SPACE accelerates inference speed, it is imperative to explore whether LLMs trained with SAR-SFT suffer performance degradation compared to those trained with the conventional SFT approach. To this end, we train LLMs with SFT under the same datasets and training  configuration used for SAR-SFT. Note that by setting $p_{\mathrm{ar}}=1$, SAR-SFT effectively becomes equivalent to SFT. 

\begin{figure}[htb]
\centering
\includegraphics[width=3in]{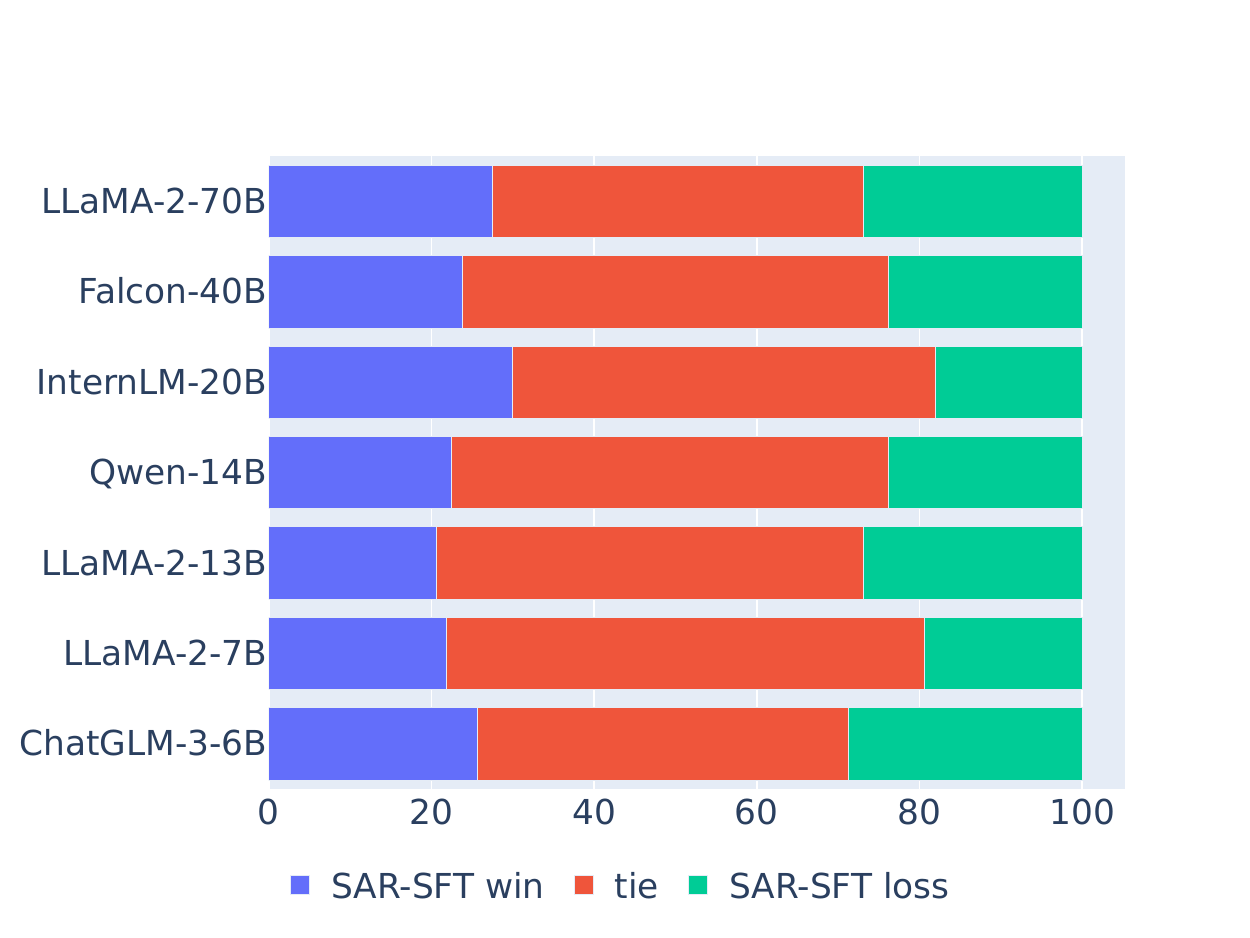}
\caption{Win rate comparison in MT-Bench: SAR-SFT versus SFT judged by GPT-4. Best viewed in color.}
\label{fig:gpt4_eval}
\end{figure}

\begin{figure}[htb]
\centering
\includegraphics[width=3in]{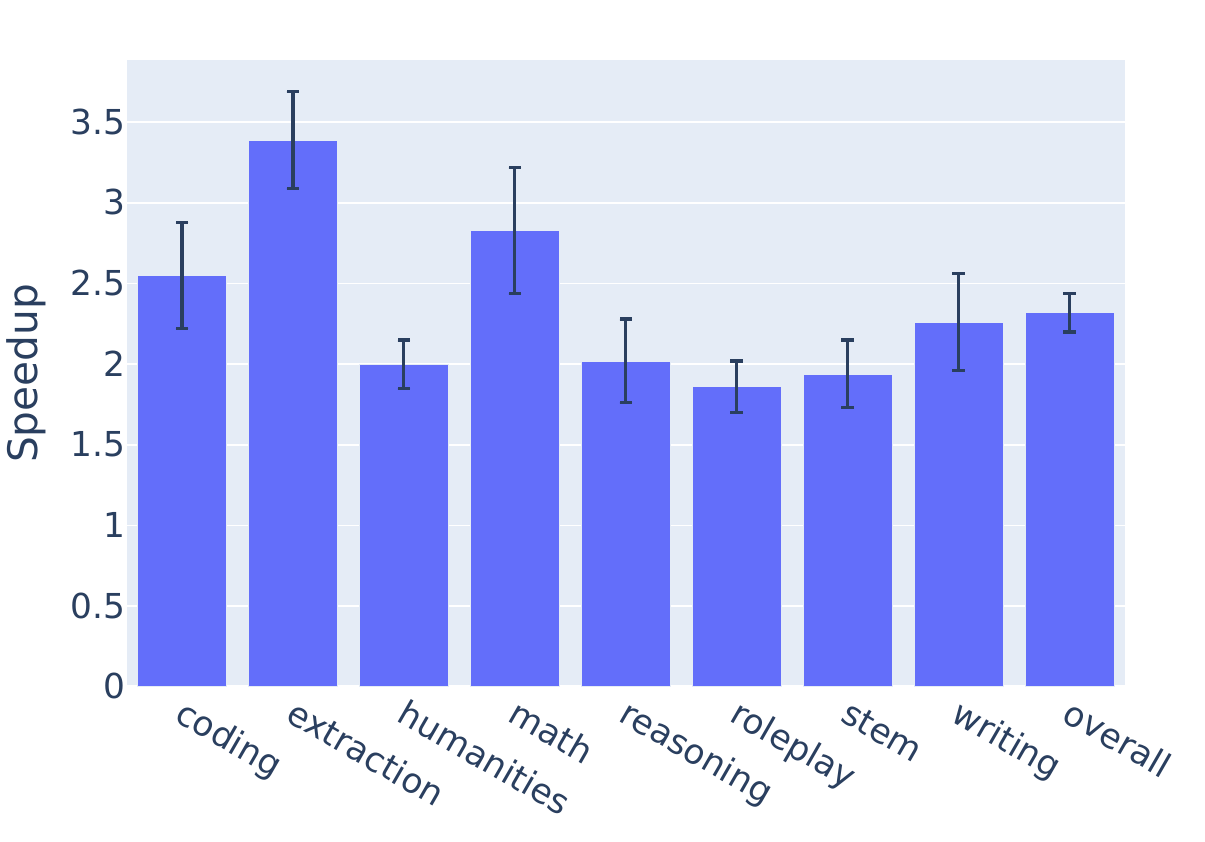}
\caption{The mean and standard deviation of speedup for all models under greedy sampling setting in MT-Bench.}
\label{fig:mt_bench}
\end{figure}

For a comprehensive comparison, MT-Bench was employed with GPT-4 serving as the evaluator to measure the performance disparity between the LLMs trained with the two training schemes. The results are presented in Figure~\ref{fig:gpt4_eval}. We can observe that models trained with SAR-SFT scheme have comparable performance as compared to their SFT counterparts. Specifically, the majority of questions assessed in MT-Bench ended in a deadlock across all models, implying that training an LLM with SAR-SFT does not deteriorate the model's quality. Additionally, SAR-SFT-trained models have exhibited advantages in speed. The mean and standard deviation of the speedup for all models in various tasks within MT-Bench are shown in Figure~\ref{fig:mt_bench}. It becomes evident that the speedup ratios vary considerably across different tasks, with the highest gains observed in tasks related to extraction, math, and coding. On average, all the models achieved a speedup ratio of 2.3 in MT-Bench dataset. More details can be found in Table~\ref{tab:result_mt_bench} in the appendix.

To further validate that SAR-SFT does not compromise the model's effectiveness, a comprehensive evaluation was conducted using a suite of widely adopted benchmarks, including MMLU, BoolQ, and others. More detailed can be found in Appendix~\ref{app:sft}.
%we undertook an exhaustive comparison of models fine-tuned using both SAR-SFT and SFT. Our assessment utilized a set of established benchmarks, including MMLU, BoolQ, and others.  Additional details are provided in Appendix~\ref{app:sft}.

\subsubsection{Ablation Study}

\begin{figure}[thb]
\centering
\includegraphics[width=3in]{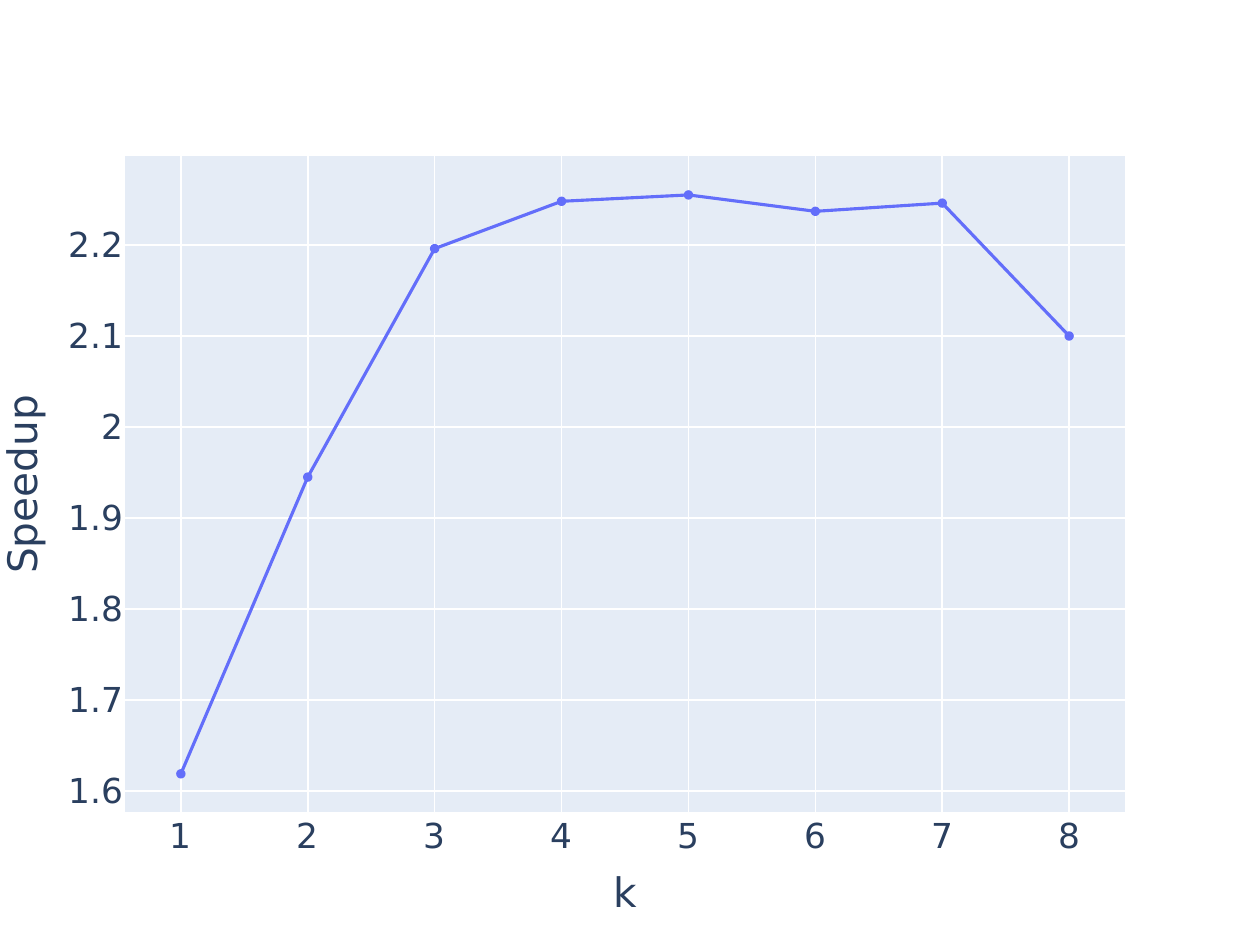}
\caption{Ablation study on number of mask tokens based on LLaMA-2-7B. The speedup are evaluated under greedy sampling setting on MT-Bench dataset.}
\label{fig:ablation_k}
\end{figure}

Our ablation study investigates the impact of varying the number of masked tokens, denoted as $k$, on the speedup ratio of the LLaMA-2-7B model using the MT-Bench dataset. The results of this analysis are presented in Figure~\ref{fig:ablation_k}. Our findings indicate that a setting of $k=5$ achieves an optimal balance for the model's performance. During the SAR-SFT phase, the LLM is tasked with concurrently predicting a sequence of $k$ subsequent tokens. Increasing the value of $k$ elevates the complexity of the prediction task and introduces computational overhead during inference, which may inversely correlate with the acceleration of the decoding process. Conversely, setting too low a value for $k$ leads to an underutilization of the model's capacity for parallel decoding, potentially resulting in a less pronounced improvement in decoding speed.

\subsection{Integration with TGI}

When deploying LLMs for production use, it's common to leverage advanced LLM serving engines designed to enhance the efficiency of text generation tasks. The Text Generation Inference (TGI)~\cite{tgi} framework is one such example, widely recognized for its support for a suite of acceleration techniques such as flash attention, tensor parallelism, and continuous batching.

\begin{figure}[!htb]
\centering
\includegraphics[width=3in]{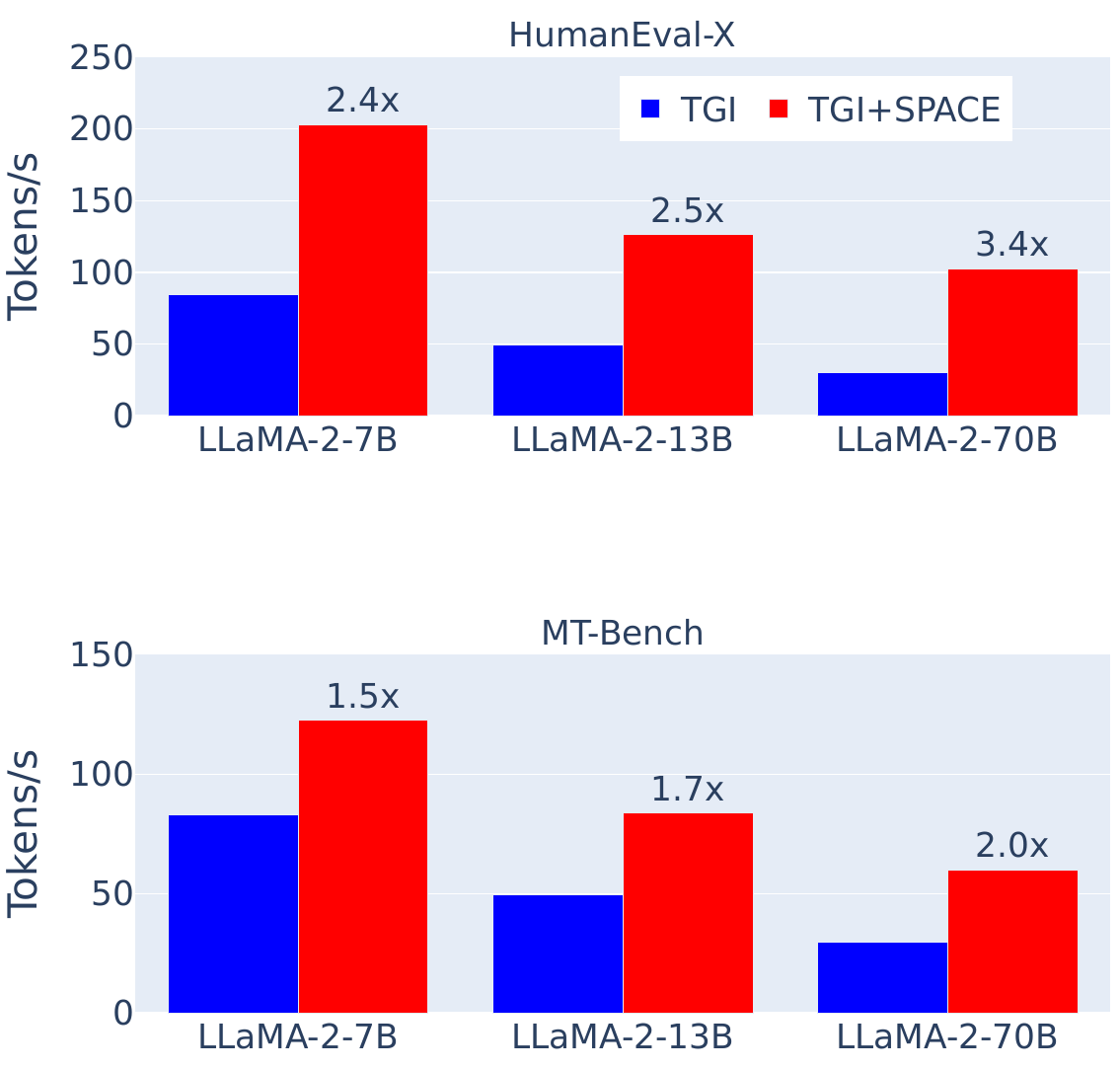}
\caption{Token generation speed (Tokens/s) and speedup for LLaMA-2 (7B, 13B, 70B)  with TGI and SPACE integration on HumanEval-X and MT-Bench datasets under greedy sampling setting.}
\label{fig:tgi}
\end{figure}

We have integrated SPACE with the TGI framework. The primary objective of this integration is to ascertain whether SPACE can yield acceleration gains even when combined with other advanced inference-optimizing techniques presented in TGI. The results shown in Figure~\ref{fig:tgi} were encouraging: with SPACE, TGI achieved a speed increase ranging from 1.5x to 3.4x across various model sizes. Remarkably, the incorporation of SPACE enabled LLaMA-2-13B model to reach inference speeds comparable to, if not surpassing, those of a 7 billion-parameter model without SPACE supports. %We will release our implementation of TGI with SPACE once the paper is accepted.

% \begin{figure}[htb]
% \centering
% \includegraphics[width=3in]{fig/tgi_human_eval.pdf}
% \caption{Speedup of different models on HumanEval-X and MT-Bench.}
% \label{fig:ablation_k}
% \end{figure}

% \begin{table*}[t]
% \centering
% \begin{tabular}{l | c | c | c | c | c }
% \hline
% \textbf{Method} &\textbf{1} & \textbf{2}& \textbf{4}& \textbf{8}& \textbf{16}  \\
% \hline
% AR & 23.8 & 22.8 & 22.1 & 21.1  & 19.5  \\
% SPACE (k=5) & 63.0 (2.64) & 56.0 (2.46) & 49.2 (2.23) & 38.4 (1.82) & 27.2 (1.39) \\
% SPACE (k=2) & 53.8 (2.26) & 51.8 (2.27) & 48.3 (2.19) & 41.4 (1.96) & 33.9 (1.74) \\
% \hline
% \end{tabular}
% \caption{The inference speed in tokens/s and speedup for AR and SPACE with different batch sizes. The number in parentheses shows the speedup.}
% \label{tab:result_batch}
% \end{table*}

To assess SPACE's efficacy with larger batch sizes, we carried out experiments on the MT-Bench dataset using the LLaMA-2-70B model through TGI. 
The results in Table~\ref{tab:result_batch} suggest a reduced speedup from SPACE as batch sizes grow. Notablely, SPACE with five masks ($k=5$) achieves only a 1.39x improvement with a batch size of 16. This diminishing speedup is due to the computational overhead introduced by the additional tokens during the inference phase, an effect that is magnified as batch size escalates. Additionally, our findings indicate that decreasing the number of masks enhances SPACE performance. Specifically with batches over 8, SPACE with two masks are more efficient than those with five. This is intuitive as a smaller value of $k$ introduces fewer additional tokens, thereby saving computational resources.
More comparisons on AR and SPACE with large batch size can be found in Appendix~\ref{app:batch_size}.

\begin{table}[t]
\centering
\begin{tabular}{l | c | c | c | c }
\hline
\multirow{2}{*}{\textbf{Method}} & \multicolumn{4}{c }{\textbf{Batch Size}} \\
\cline{2-5}
&\textbf{2}& \textbf{4}& \textbf{8}& \textbf{16}  \\
\hline
AR & 22.8 & 22.1 & 21.1  & 19.5  \\
\hline
\makecell[c]{SPACE\\(k=5)}  & \makecell[c]{56.0\\ (2.46)} & \makecell[c]{49.2\\ (2.23)} & \makecell[c]{38.4\\ (1.82)} & \makecell[c]{27.2\\ (1.39)} \\
\hline
\makecell[c]{SPACE\\ (k=2)} & \makecell[c]{51.8\\ (2.27)} & \makecell[c]{48.3\\ (2.19)} & \makecell[c]{41.4\\ (1.96)} & \makecell[c]{33.9\\ (1.74)} \\
\hline
\end{tabular}
\caption{The inference speed in tokens/s per request and speedup for AR and SPACE on MT-Bench dataset. The number in parentheses shows the speedup.}
\label{tab:result_batch}
\end{table}

\section{Conclusion}
%In this paper, we introduce SPACE, an innovative approach to accelerate the inference speed of large language models. 
In this paper, we introduce SPACE, an innovative approach to accelerate inference of LLMs. SPACE is distinguished by 1) its ability to transform an AR LLM into a SAR LLM utilizing SAR-SFT, which is easy to implement as it only requires  minor modifications to the dataloader in SFT setup;
2) its unique auto-correct decoding algorithm that enables the same model for both token generation and verification.
%SPACE seamlessly incorporates a semi-autoregressive model with a novel draft-then-verify inference algorithm. 
%It achieves this by implementing a tailored semi-autoregressive supervised fine-tuning procedure that enhances the capabilities of current LLMs, enabling them to predict several tokens concurrently. Moreover, SPACE employs an innovative auto-correct decoding strategy that allows for the concurrent generation and verification of token sequences during a single invocation of the model.
%Our experiments reveal that an autoregressive LLM, fine-tuned in a semi-autoregressive approach, can generate likely sequences of tokens in parallel. The adoption of an effective auto-correct decoding algorithm facilitates the simultaneous generation and verification of token sequences. 
Experimental results on various LLMs show SPACE can achieve 2.7x-4.0x speedup on HumanEval-X while still preserving model quality. 
%This speedup is on par with, if not surpassing, alternative approaches such as speculative decoding, demonstrating SPACE as a competitive technique for enhancing inference efficiency in LLMs.
% \section*{Acknowledgments}
% This was was supported in part by......

\section{Limitations}
While SPACE has demonstrated potential in accelerating the inference of LLMs, it also brings about certain limitations that must be acknowledged: First, the primary advantage offered by SPACE is the acceleration of the inference process through the introduction of additional input tokens during decoding, which has the potential to reduce the number of forward passes that LLMs require. 
%However, this approach does not completely address the concern of computational overhead. 
%The presence of these additional tokens can, in some instances, lead to an increase in computational load, potentially diminishing the efficiency gains offered by the decreased number of forward passes. 
However, the presence of these additional tokens inevitably leads to increased computation overhead, notably in terms of FLOPs, when compared to conventional autoregressive decoding.
Therefore, it becomes crucial to conduct an exhaustive study on the energy consumption of methods like SPACE, to fully understand and mitigate their ecological impact. The sustainability of deploying such acceleration techniques, considering long-term environmental implications, must factor into the development of responsible AI technologies.
%Our method introduces an augmentation to the input token sequence by inserting $k*(k+2)$ supplementary tokens at each decoding step, which inevitably leads to increased computation overhead, notably in terms of FLOPs, when compared to conventional autoregressive decoding. It is worth noting, however, that during the inference phase of an LLM, the bottleneck is often the memory bandwidth. In other words, the inference speed is constrained by the rate at which model parameters are transferred from the GPU memory to local registers. The actual computation on the data once it's loaded tends to be less of a limiting factor. Additionally, our approach benefits from the parallel processing of the added tokens. Therefore, for a reasonably chosen value of $k$, the impact of these additional tokens on the inference speed turns out to be negligible.
%This heightened power requirement could translate into greater energy consumption, a notable concern from an environmental standpoint. Thus a comprehensive study in terms of energy consumption of SPACE need to be conducted, and the additional overhead introduced by SPACE must be considered in the context of sustainability and the long-term implications of deploying such methods.

Furthermore, it is important to recognize that the gain in inference speed facilitated by SPACE is variable across different tasks. Our empirical observations suggest that the speedup is inconsistent, and the limited datasets examined in this study could contribute to skewed outcomes. Besides, our evaluations for SPACE were conducted exclusively on English datasets; consequently, the extent to which SPACE can accelerate inference in other languages has not yet been investigated. It is plausible that there are specific datasets where SPACE exhibits a significantly lower degree of acceleration—a scenario not captured within the confines of our experimental array. 

%Moreover, we do not compare SPACE with other inference-accelerating methods in this paper. The lack of a standardized benchmark combined with the potential variability introduced by different model architectures, evaluation datasets, and hardware configurations makes such comparisons challenging. Rather than drawing indirect comparisons based on the speedup ratios reported from previous work, we aim to provide a more equitable evaluation by reproducing selected existing methods and assessing them using an identical setup in our future work.

Lastly, we leveraged MT-Bench along with a collection of well-established benchmarks, such as MMLU, PIQA, AGIEval, and others, to gauge model performance when trained with SAR-SFT as opposed to traditional SFT methodologies. Despite this extensive set of evaluations, it is critical to emphasize that benchmarking the comprehensive capabilities of LLMs remains a challenge, and the datasets engaged in this research fall short of enabling a definitive judgment. To this end, we advocate for the application of SPACE in diverse downstream tasks by the research community, which will offer a more rounded understanding of its practical utility and limitations.

%Besides, the additional tokens and specially designed attention mask introduce implementation difficulty, more engineering effort.

% Entries for the entire Anthology, followed by custom entries
\bibliography{anthology,custom}
\bibliographystyle{acl_natbib}

\appendix
\section{Appendix}

\subsection{Training Details}
\label{app:table}

We conduct all our experiments on a cluster of 4 servers, where each server is equipped with eight A800 (80G) GPUs. We adopt distinct training strategies based on the size of the models being trained. For models with fewer than 14 billion parameters, we allocate our experiments to a single server and employ the ZeRO-2~\cite{rasley2020deepspeed} optimization for distributed training. Conversely, for models that exceed the 14 billion parameter mark, we expand our setup to utilize all four servers and implement the ZeRO-3 optimization to effectively handle the increased computational demands. We adopt LLaMA Factory~\cite{llama-factory} to fine-tune the LLMs. The specific hyper-parameters utilized for the SAR-SFT are documented and can be referenced in Table~\ref{tab:training_conf}. 
%Typically, it costs around 6 hours to finetune the LLaMA-2-7B on a server with eight A800 (80GB) GPUs. For the largest model, i.e.,  LLaMA-2-70B, the SAR-SFT costs around 18 hours on 4 servers with 32 A800 (80GB) GPUs.

\begin{table}[htb]
\centering
\begin{tabular}{c c}
\hline
\textbf{Hyper-parameters} & \textbf{Value} \\
\hline
max source tokens & 2048 \\
max target tokens & 2048 \\
learning rate & 5e-5\\
scheduler & cosine \\
Adam $\beta_1$ & 0.9 \\
Adam $\beta_2$ & 0.999 \\
epoch & 2 \\
per device batch size & 4 \\
gradient clip & 1.0 \\
\hline
\end{tabular}
\caption{Hyper-parameters and training configurations of SAR-SFT.}
\label{tab:training_conf}
\end{table}

Table~\ref{tab:dataset} shows the statistics of SFT datasets used to finetuned the models. Note that all the dataset are publicly available. The fine-tuning duration for LLMs can vary significantly based on the size of the model and the computational resources available. For the LLaMA-2-7B model, the fine-tuning process typically takes about 6 hours on a server equipped with eight A800 (80GB) GPUs. For the largest variant, the LLaMA-2-70B, the SAR-SFT requires roughly 18 hours to complete using 4 servers, each equipped with 8 A800 (80GB) GPUs (totalling 32 GPUs). 
%In regards to the experimental duration, we observed that training the LLaMA-2-7B model typically required approximately six hours on a server configured with eight A800 (80G) GPUs. 

\begin{table*}[!htb]
\centering
\begin{tabular}{l c c c c}
\hline
\textbf{Dataset} & \textbf{Language} & \makecell[c]{\textbf{Sample}\\ \textbf{Numbers}}  & \makecell[c]{\textbf{Average}\\ \textbf{Input Tokens}} & \makecell[c]{\textbf{Average}\\ \textbf{Output Tokens}} \\
\hline
Alpaca-GPT4-zh~\cite{peng2023instruction} & zh & 48,818 & 30.9 & 292.5 \\
Alpaca-GPT4-en~\cite{peng2023instruction}  & en & 52,002 & 21.6 & 162.6 \\ LIMA~\cite{zhou2023lima} & en & 1,029 & 74.2 & 639.1 \\ 
Oaast-SFT~\cite{OpenAssistant} & multi & 20,202 & 198.8 & 234.8 \\ CodeAlpaca~\cite{codealpaca} & en & 20,022 & 28.8 & 68.6 \\ OpenPlatypus~\cite{platypus2023} & en & 24,926 & 159.6 & 225.3 \\
\hline
\end{tabular}
\caption{Statistics of SFT datasets used to finetuned the models. The average input tokens and output tokens are calculated using LLaMA-2-7B tokenizer.}
\label{tab:dataset}
\end{table*}

We observe that the introduction of SAR-SFT imposes negligible additional training costs when compared to conventional SFT. This is largely due to the minimal intervention in the training process: we only inject $k=5$ additional mask tokens during SAR-SFT training. Specifically, the adaptation concerns solely the dataloader—other components are kept consistent with the standard SFT approach.
In this modified dataloader, each data sample remains unchanged with a probability $p_{\mathrm{ar}}$. Conversely, with a probability of $1-p_{\mathrm{ar}}$, we randomly select a position $m$ in the input sequence to replace $k$ consecutive tokens with mask tokens, while the label sequence is left intact. We then truncate the input and label token sequences to keep the first $m+k$ tokens. 
%This approach incurs minimal overhead, allowing us to maintain training efficiency. 
Compared to the standard SFT, which does not employ truncation on the training sample, the training cost of SAR-SFT is marginally reduced, as tokens beyond the positions $m+k$ are discarded during the training process in SAR-SFT. Namely, SAR-SFT actually sees less tokens during training than SFT.
Empirical evidence from our experiments supports this assertion, as we recorded similar training durations for both SAR-SFT and SFT under identical training configurations. 

\subsection{Evaluation Details}
\label{app:eval}
We performed our inference experiments on a server equipepd with eight A800 (80GB) GPUs. For models with fewer than 14 billion parameters, inference is conducted using a single GPU. For larger models, those with parameters exceeding 14B, we employ multiple GPUs and leverage tensor parallelism to manage the increased computational load effectively. During the inference process, we configure our setup with a batch size of one to ensure precise measurement of inference latency on a per-instance basis.

In our experiment, we employ four distinct datasets: Chatbot Instruction Prompt (CIP) \cite{cip2023}, MT-Bench \cite{zheng2023judging}, HumanEval-X \cite{zheng2023codegeex} and XSum \cite{narayan2018don}. CIP is a conversational dataset from which we utilize prompts to simulate realistic conversations. MT-Bench is a dataset comprised of multi-turn questions, encompassing a wide range of topics. 
HumanEval-X is a standard benchmark for Python code generation and Pass@10 is used as the metric.
Lastly, the XSum dataset, which tasks models with summary generation, is evaluated using  ROUGE-L. 

For generation tasks, we tailored specific prompt templates to guide the model's output. When working with the XSum dataset, we used the following prompt template:``Document: \{TEXT\}\textbackslash n Based on the previous text, provide a brief single summary''. Similarly, for the HumanEval-X dataset, which is designed for code generation, we employed the prompt template as follows: ``Complete the following python code. Do not give any explanation or testing examples, just complete the code.\textbackslash n \{TEXT\}''. For CIP and MT-Bench, we do not use any prompt template.

To mesure the performance of LLMs on XSum and HumanEval-X, we compute the ROUGE-L and Pass@10, respectively. The ROUGE-L is calculated using python package rouge~\cite{rouge} and the pass@10 is computed using official evaluation script~\cite{zheng2023codegeex}.

The inference speedup for each task within the MT-Bench benchmark under greedy sampling setting across various models are shown in Table~\ref{tab:result_mt_bench}.

\begin{table*}[htp]
\centering
\begin{tabular}{r | c | c | c | c | c | c | c | c | c}
\hline
\textbf{Model} & \textbf{Code} & \makecell[c]{\textbf{Extrac-}\\\textbf{tion}} & \makecell[c]{\textbf{Human-}\\\textbf{ities}} & \textbf{Math} & \makecell[c]{\textbf{Reason-}\\\textbf{ing}} & \makecell[c]{\textbf{Role-}\\\textbf{play}} & \textbf{Stem} &\makecell[c]{\textbf{Writ-}\\\textbf{ing}} & \makecell[c]{\textbf{Over-}\\\textbf{all}}\\
\hline
ChatGLM-3-6B  & 2.83 & 3.35 &1.91 &2.54 & 1.87 & 1.98 & 2.03 & 2.43 & 2.32 \\
LLaMA-2-7B &  2.14 & 3.12 & 1.96 & 2.61 & 1.83 & 1.89 &1.65 & 2.25 &2.19\\
LLaMA-2-13B&2.89  & 3.66 & 2.20 & 2.99  & 2.29 & 2.03 & 2.27 & 2.68 & 2.53\\
Qwen-14B & 2.88  & 3.76 & 2.18 & 2.85 & 2.04 & 1.86 & 2.05 & 2.55 &2.43 \\
InternLM-20B & 2.50  & 3.28 & 2.05 & 3.55  & 1.89 & 2.00 & 2.02 & 2.05 &2.36 \\
Falcon-40B & 2.02  & 2.90 & 1.72 & 2.22 & 1.69 & 1.53 & 1.69 & 1.76 & 2.17 \\
LLaMA-2-70B & 2.61  & 3.70 & 1.97 & 3.03  & 2.50 & 1.73 & 1.84 & 2.09 & 2.26 \\
\hline
\end{tabular}
\caption{The experimental results on MT-Bench under greedy sampling setting. We show the inference speedup for each task in MT-Bench.}
\label{tab:result_mt_bench}
\end{table*}

\subsection{Attention Mask in SPACE}
\label{app:attention}

\begin{figure}[!htb]
\centering
\includegraphics[width=3in]{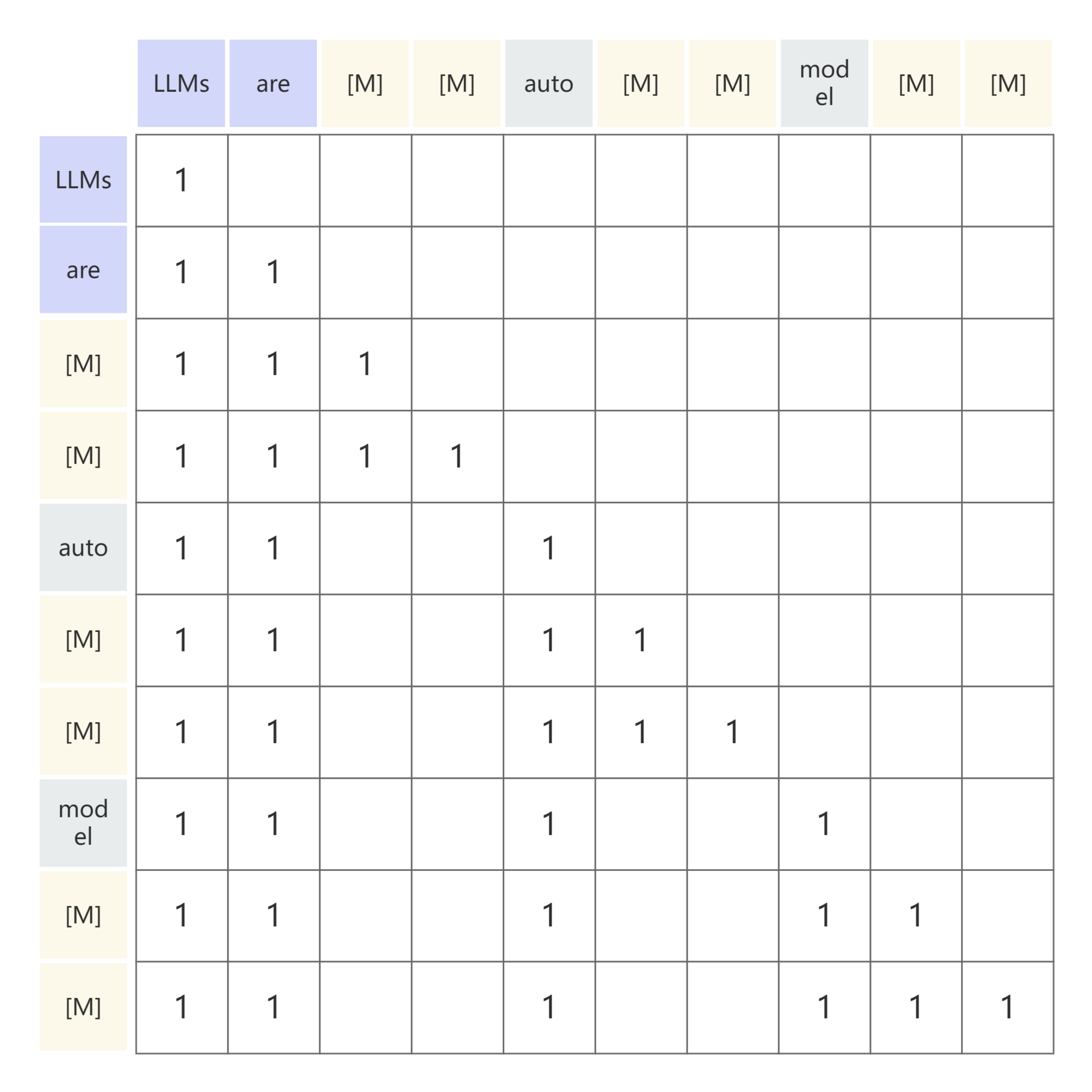}
\caption{An illustrative example of the attention mask used in SPACE. In this example, $k=2$ and the input is extended with $8$ tokens. ``LLMs are'' are the input query, ``auto'' and ``model'' are two candidate tokens that need to be verified.}
\label{fig:llm_attention}
\end{figure}
An illustrative example of the attention mask is shown in Figure~\ref{fig:llm_attention}.
The attention mask used in SPACE is tailored such that masked tokens can causally attend only to other mask tokens within the same group and to preceding non-masked tokens. Furthermore, all non-masked tokens are restricted to causally attend to prior non-masked tokens, and are unable to attend to any preceding masked tokens.

\subsection{Random Sampling}
\label{app:random_sampling}
%Similar observations can be found when random sampling is enable, as shown in Table~\ref{tab:result_xsum_humaneval}.
%We evaluate the models on XSum and HumanEval-X when random sampling is enabled with parameters top-p=0.95 and top-k=10. To eliminate impact of randomness, we run the evaluation for 10 times and report the mean and variance in Table~\ref{tab:result_xsum_humaneval}. With random sampling, the output of SPACE and baseline are expected to follow the same distribution. This is verified in the table that the performance in both XSum and HumanEval-X between SPACE and baseline are almost the same.
To rigorously evaluate model performance on the XSum and HumanEval-X datasets with random sampling enabled~\footnote{When using random sampling, we set top-p=0.95 and top-k=10}, we conducted ten runs of the evaluation process to counteract the influence of randomness. The mean and variance of these runs are reported in Table~\ref{tab:result_xsum_humaneval}. Under random sampling setting, the performance metrics for SPACE and the baseline are similar on both XSum and HumanEval-X, as presented in Table~\ref{tab:result_xsum_humaneval}. This consistency across multiple evaluations confirms the distributional alignment between SPACE and the baseline model under the random sampling setting.

\begin{table*}[!ht]
\centering
\begin{tabular}{r | c | c | c | c | c | c }
\hline
\multirow{2}{*}{\textbf{Model}} & \multicolumn{3}{c |}{\textbf{XSum}} & \multicolumn{3}{c}{\textbf{HumanEval-X}} \\
\cline{2-7}
& ROUGE-L & \makecell[c]{Avg. \\ Tokens}  & Speedup & Pass@10 & \makecell[c]{Avg. \\ Tokens}  &  Speedup \\
\hline
ChatGLM-3-6B  & \makecell[c]{$14.8 \pm 0.2$\\($14.0\pm 0.4$)} & $1.95\pm 0.01$ & $1.47\pm 0.01$ & \makecell[c]{23.2 \\ (22.8)} & $3.16\pm 0.04$ & $2.09 \pm 0.08$ \\
\hline
LLaMA-2-7B & \makecell[c]{$15.1\pm 0.2$ \\ ($15.3\pm 0.1$)} & $2.14\pm 0.02$ & $1.79\pm 0.04$ & \makecell[c]{18.9 \\ (18.3)} & $3.56\pm 0.05$ & $2.86\pm 0.04$ \\
\hline
LLaMA-2-13B & \makecell[c]{$15.2\pm 0.2$\\($15.6\pm 0.2$ )} & $2.24\pm 0.01$ & $1.86\pm 0.02$ & \makecell[c]{31.7 \\ (32.3)} & $4.15 \pm 0.02$ & $3.81 \pm 0.05$ \\
\hline
Qwen-14B & \makecell[c]{$16.1\pm 0.3$\\ ($16.3\pm 0.3$)} & $2.05\pm 0.01$ & $1.91\pm 0.04$ & \makecell[c]{32.3 \\ (31.7)} & $3.09\pm 0.04$ & $2.86\pm 0.04$ \\
\hline
InternLM-20B & \makecell[c]{$16.3\pm 0.2$ \\ ($17.0\pm 0.2$)} & $1.99 \pm 0.01$ & $1.73\pm 0.01$ & \makecell[c]{25.0 \\ (23.7)} & $3.13\pm 0.03$ & $2.67\pm 0.08$\\
\hline
Falcon-40B & \makecell[c]{$16.6\pm 0.2$\\($15.4\pm 0.3$)} & $2.09\pm 0.04$ & $2.08\pm 0.03$ & \makecell[c]{27.4 \\ (28.0)} & $3.42\pm 0.03$ & $2.88\pm 0.06$\\
\hline
LLaMA-2-70B & \makecell[c]{$16.1\pm 0.2$\\($16.2\pm 0.3$)} & $2.40\pm 0.02$ & $2.25\pm 0.02$ & \makecell[c]{36.6 \\ (38.2)} & $4.15\pm 0.02$ & $3.81\pm 0.05$ \\
\hline
\end{tabular}
\caption{The experimental results on XSum and HumanEval-X using random sampling. We show the mean and variance (over 10 runs) of the average accepted tokens (Avg. Tokens) and inference speedup (Speedup) for each datasets. The number in parentheses shows the corresponding results of the baseline method.}
\label{tab:result_xsum_humaneval}
\end{table*}

\subsection{SAR-SFT versus SFT}
\label{app:sft}
To further demonstrate that SAR-SFT does not impede the model's performance, we compared the performance of LLaMA-2 (with model sizes of 7B, 13B, and 70B parameters) trained with both SAR-SFT and traditional SFT. The comparison spanned a suite of widely used benchmarks, which we have categorized into the following four groups:
\begin{itemize}
    \item \textbf{Academic}. We report the average accuracy of the model on the MMLU~\cite{hendrycks2020measuring} and AGIEval~\cite{zhong2023agieval} benchmarks.
    \item \textbf{Knowledge}. We evaluate the model on CommonSenseQA~\cite{talmor2018commonsenseqa} and BoolQ~\cite{clark2019boolq}, reporting their average results. 
    \item \textbf{Reasoning}. We assess the 5-shot performance on PIQA~\cite{bisk2020piqa}, RTE~\cite{bentivogli2009fifth} and HellaSwag~\cite{zellers2019hellaswag}, reporting their mean performance.
    \item \textbf{Understanding}. We report the average result on RACE~\cite{lai2017large} and SQuAD2.0~\cite{rajpurkar2018know}.
\end{itemize}
%We conduct the evaluations using opencompass, a platform for large model evaluation~\cite{2023opencompass}. The comparative performance results are detailed in Table~\ref{tab:performance}. Upon examination, we note negligible discrepancies between the models finetuned using two distinct training schemes SAR-SFT and SFT. 
The evaluations were conducted using OpenCompass~\cite{2023opencompass}, an opensource platform designed for large language model evaluation. Comparative performance results are detailed in Table~\ref{tab:performance}. Upon examination of the results, we note small discrepancies between the models fine-tuned with the two distinct training schemes across different tasks.

%This similarity in performance could be attributed to two potential factors: The default setting of the autoregressive probability ($p_{\mathrm{ar}}$) to 0.5 implies that the objectives of SAR-SFT and SFT coincide half of the time, which may result in minimal divergences between the two schemes.

\begin{table*}[!htb]
\centering
\begin{tabular}{r | c | c | c | c | c}
\hline
\textbf{Model} & \textbf{Scheme} & \textbf{Academic}  &  \textbf{Knowledge} & \textbf{Reasoning} & \textbf{Understanding} \\
\hline
\multirow{2}{*}{LLaMA-2-7B} & SAR-SFT & 35.4 & 66.1 & 62.3 &  37.2\\
& SFT  & 36.0 & 65.9 & 64.1 & 38.6 \\
\hline
\multirow{2}{*}{LLaMA-2-13B} &  SAR-SFT & 40.9 & 69.4 & 66.7 & 55.2\\
& SFT  &  40.5 & 71.4 & 65.2 & 57.4\\
\hline
\multirow{2}{*}{LLaMA-2-70B} & SAR-SFT & 50.6 & 76.7 & 68.4 &  64.7\\
 & SFT  & 51.7 & 77.2 & 68.0 &  66.7\\
\hline
\end{tabular}
\caption{Performance comparison of LLaMA-2 (7B, 13B, 70B) with different training schemes.}
\label{tab:performance}
\end{table*}

% \subsection{Clarify of Contribution}
% \label{app:sft}

\subsection{Discussion on Large Batch Size}
\label{app:batch_size}

\begin{figure}[!htb]
\centering
\includegraphics[width=3in]{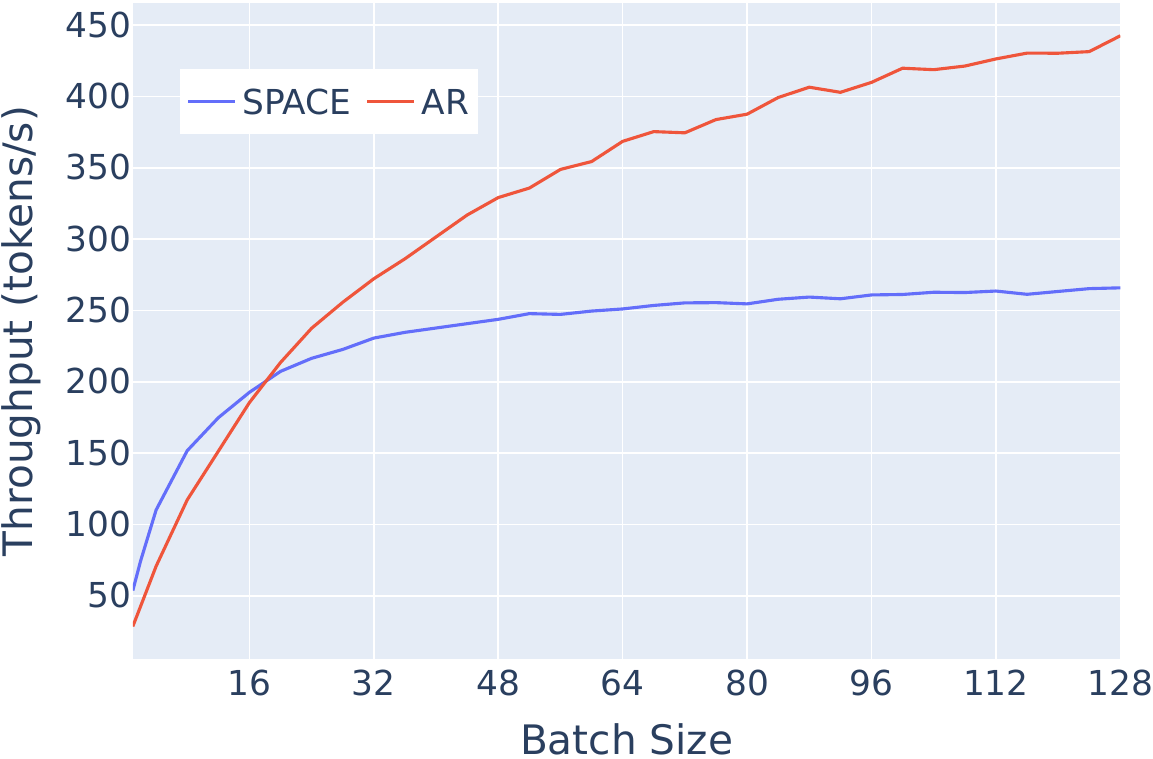}
\caption{Throughput of SPACE and AR under various batch sizes on XSum dataset.}
\label{fig:llm_throughput}
\end{figure}

When operating at the same batch sizes, SPACE consumes more GPU computational resources to speed up inference in comparison to AR. Nevertheless, with an increase in batch size, the GPU becomes compute-bound, effectively neutralizing SPACE's performance advantage. To provide a fair assessment of SPACE and AR, we perform experiments on the XSum dataset for both methods. Our focus is on evaluating their throughput, defined as the number of output tokens per second that an inference server can produce across all requests, while experimenting with various batch sizes. The findings, illustrated in Figure~\ref{fig:llm_throughput}, reveal that SPACE's throughput significantly diminishes when compared to AR at larger batch sizes. Specifically, the throughput for SPACE plateaus when the batch size reaches approximately 64, whereas AR's throughput does not saturate until around a batch size of 128. However, SPACE remains competitive at lower batch sizes. Notably, with batch sizes smaller than 16, SPACE's throughput exceeds that of AR.

% To make a fair comparison between SPACE and AR under consistent GPU usage, we document the shortest end-to-end inference times on the XSum dataset for both methods. This is accomplished by incrementally increasing the batch size until we identify the minimum inference time. The results are shown in Table~\ref{tab:result_batch_fastest}. We observe that SPACE (k=5) outperforms AR with a minimum inference time of XXXs at a batch size of XXX on a mahcine with 8*A800 80GB GPUs.

% \begin{table}[htp]
% \centering
% \begin{tabular}{l | c | c | c}
% \hline
% \textbf{Method} & \textbf{GPUs} & \textbf{Time} &\textbf{Batch Size} \\
% \hline
% AR & 8*A800 & 136.5s & 35\\
% SPACE & 8*A800 & 116.8s & 31\\
% AR & 8*4090 & \\
% SPACE & 8*4090 & 570.3s & 20\\
% AR & 8*3090 &\\
% SPACE & 8*3090 &\\
% \hline
% \end{tabular}
% \caption{Minimum achievable inference time of AR and SPACE with LLaMA-2-70B on XSum dataset.}
% \label{tab:result_batch_fastest}
% \end{table}

Furthermore, it is worth noting that many practical applications, especially those on edge devices, often operate with a small batch size. In such cases, the inference process remains memory-bound rather than compute-bound. SPACE can effectively utilize the available computing resources to enhance the inference speed in low batch size scenarios, making it an attractive and valuable solution for such use cases.
\end{document}